\documentclass{article}

\usepackage[dblblindworkshop,final]{neurips_2026}
\workshoptitle{Long-Context Foundation Models}

\usepackage[utf8]{inputenc}
\usepackage[T1]{fontenc}
\usepackage{hyperref}
\usepackage{url}
\usepackage{booktabs}
\usepackage{amsfonts}
\usepackage{amsmath}
\usepackage{nicefrac}
\usepackage{microtype}
\usepackage{xcolor}
\usepackage{graphicx}
\usepackage{multirow}
\usepackage{subcaption}
\usepackage{enumitem}
\usepackage{seqsplit}
\usepackage{tikz}
\usetikzlibrary{positioning,arrows.meta,fit,backgrounds,calc}
\usepackage{cleveref}
\newcommand{\methodname}{Multi-layer Indexer}
\newcommand{\MLI}{MLI}
\newcommand{\IC}{IndexCache}

\newcommand{\Isel}{\mathcal{S}}

\newcommand{\relu}{\mathrm{ReLU}}
\newcommand{\ret}[1]{\ensuremath{\nicefrac{1}{#1}}}

\title{\methodname: Overlapping Selection\\and Attention in DeepSeek Sparse Attention}

\author{%
  Serdarcan Dilbaz, Arjun Krishna, Kiran Kamble, Daniel M. Bikel \\
  Writer AI Research, Writer, Inc. \\
  \texttt{\{serdarcan,arjun,kiran,dan.bikel\}@writer.com}
}

\begin{document}

\maketitle

\begin{abstract}
Sparse attention cuts the cost of long-context inference by making the attention matrix sparse: each query attends only a subset of the sequence, chosen by a fixed pattern or, in recent production LLMs, learned per query. In DeepSeek Sparse Attention (DSA), the attention of DeepSeek-V3.2 and the GLM-5 family, every layer is preceded by a \emph{lightning indexer} that scores the whole cache and keeps the $k{=}2048$ highest-scoring tokens for the layer to attend. At long context this indexer scan, not the sparse attention it feeds, is where the decode time goes, and attention waits for it at every layer. But adjacent layers select nearly the same tokens, so a layer can attend the selection of the previous indexer layer instead of its own. This removes the dependency: each indexer then runs concurrently with the attention of its layer. The resulting \methodname{} (\MLI) needs no retraining and no new weights. Its default realization changes no kernel, running inside SGLang's production decode graph; an optional fused decode kernel accelerates it further on DeepSeek-V3.2. \MLI{} composes with IndexCache (the cross-layer reuse GLM-5.2 ships as IndexShare): IndexCache drops the indexer from some layers, and \MLI{} hides the scan on the rest. We serve DeepSeek-V3.2 and GLM-5 under IndexCache's protocol at retentions $\ret{2}$ to $\ret{32}$ (the fraction of layers that keep an indexer) and contexts $10$K to $195$K. At equal retention the composition decodes faster than IndexCache: up to $14.5\%$ faster on DeepSeek-V3.2 and $9.8\%$ on GLM-5 at $\ret{2}$, and $7.9\%$ and $5.1\%$ at $\ret{4}$, largest where IndexCache keeps the most indexers and growing with context. At retention $\ret{r}$ it is faster still than IndexCache at $\ret{2r}$ while keeping twice the fresh selections: at $128$K it reaches $1.32\times$ the unmodified model's decode throughput at $\ret{4}$, above IndexCache's $1.28\times$ at $\ret{8}$. Its LongBench, RULER and LongBench~v2 accuracy stays within a few percentage points of IndexCache's. Code is available at \url{https://github.com/ArjunKrishnak/dsa-multi-layer-indexer}.
\end{abstract}

\section{Introduction}
\label{sec:intro}

Sparse attention cuts the cost of long-context decoding by making the attention matrix sparse: each query attends only a subset of the sequence, chosen by a fixed pattern or learned per query. Recent production LLMs take the learned route, with a lightweight selector that must run before attention; at long context it, not the sparse attention it feeds, is where the decode time goes. DeepSeek Sparse Attention (DSA)~\citep{liu2025deepseek} is the attention mechanism of DeepSeek-V3.2 and of GLM-5, GLM-5.1 and GLM-5.2~\citep{zeng2026glm,zai2026glm51,zai2026glm52}. Consider decoding token $t$ with $L$ tokens in the cache. At each layer a lightning indexer with $H_I$ small heads scores every cached token $n\le L$,
\begin{equation}
  I_{t,n} \;=\; \sum_{h=1}^{H_I} w_{t,h}\, \relu\!\big(\mathbf{q}^{I}_{t,h}\cdot\mathbf{k}^{I}_{n,h}\big),
  \label{eq:indexer}
\end{equation}
where $I_{t,n}$ is the indexer's score for query $t$ and cached token $n$, the superscript $I$ marks the indexer's own projections (distinct from the layer's attention), $h$ indexes its $H_I$ heads, and $\mathbf{q}^{I}_{t,h}$, $\mathbf{k}^{I}_{n,h}$ are low-rank projections of the layer input with $w_{t,h}$ a per-head weight. The layer keeps the set $\Isel_t$ of the $k{=}2048$ tokens with the highest score and runs attention over those tokens only. The selection is the sole coupling between the indexer and attention: attention cannot begin until the top-$k$ is known, and nothing else in the layer depends on it.

\begin{figure}[t]
\centering
\includegraphics[width=0.92\linewidth]{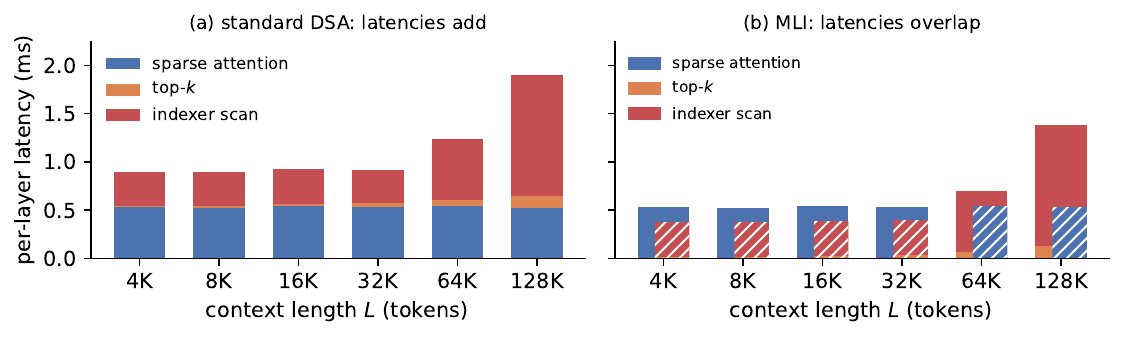}
\caption{Per-layer decode latency of the DSA attention path (DeepSeek-V3.2, H100, one request; kernels timed in isolation). \textbf{(a)} Standard DSA runs the indexer and attention in sequence, so their latencies add; the prefix scan grows with context and comes to dominate. \textbf{(b)} \MLI{} overlaps them on a second stream, so the per-layer time is the bottleneck, not the sum; the full-width bar is the bottleneck and the striped bar the hidden component, the indexer split into scan and top-$k$ as in (a). Idealized ceiling; the realized gain is smaller and bandwidth-bound (\Cref{sec:res-mechanism}).}
\label{fig:motivation}
\end{figure}

The two nevertheless run in sequence. Sparse attention gathers $k$ keys and its cost is flat in $L$; the indexer scans all $L$ keys. Beyond a few tens of thousands of tokens the scan is the largest term on the attention side of the layer, two thirds of it at $128$K (\Cref{fig:motivation}). Only the selection dependency forces the order, and adjacent DSA layers select strongly overlapping sets (\Cref{sec:background}), a cross-layer redundancy that reuse methods already exploit~\citep{bai2026indexcache,yang2025tidaldecode,deshmukh2025kascade}, so a layer's attention need not wait for its own selection: an earlier layer's serves.

The cost of the indexer per decode step is a product of two independent factors: \emph{how many} indexers run, and \emph{how long} each one is waited for. IndexCache~\citep{bai2026indexcache} reduces the first, keeping the indexer on a fraction $\ret{r}$ of the layers (\Cref{sec:background}). We reduce the second. The \methodname{} (\MLI) lets layer $\ell$ attend the selection produced by the previous indexer layer. The indexer of layer $\ell$ then feeds a later layer, not its own attention, and the two run at the same time (in the idealized limit of \Cref{fig:motivation}; the realized speedup is smaller, for reasons we trace in \Cref{sec:res-mechanism}). Every selection is still recomputed at every decode step; nothing is reused across steps. Because the two factors multiply and neither method touches the other's, the two remedies compose: on the layers IndexCache keeps, \MLI{} hides the scan that IndexCache still pays. Like IceFormer's deployable acceleration~\citep{mao2024iceformer}, \MLI{} needs no retraining, applies to every DSA model, and leaves the selections unchanged. This is the regime long context forces: KV memory caps how many requests decode together (six at $128$K, two at $195$K on our node), so very long context is served few-at-a-time, exactly where the indexer scan is exposed; the gain fades only when many short requests pack the batch and saturate memory bandwidth (\Cref{sec:res-mechanism}).

Our contributions are the following.
\begin{itemize}[leftmargin=1.4em,itemsep=1pt,topsep=2pt]
  \item \textbf{A drop-in schedule that overlaps the indexer with attention.} \MLI{} (\Cref{sec:method}) makes a layer's indexer and its attention independent, at any retention and lag depth, so they run concurrently; it needs no retraining and no new weights, runs in SGLang's decode CUDA graph with the production kernels unchanged, and is faster still on DeepSeek-V3.2 with a fused decode kernel (\Cref{sec:res-fused}).
  \item \textbf{Faster at equal retention, and one further halving for free.} Serving DeepSeek-V3.2 and GLM-5 under IndexCache's protocol (\Cref{sec:res-speed}), IndexCache$+$\MLI{} decodes faster than IndexCache at the same retention: up to $14.5\%$ faster on DeepSeek-V3.2 and $9.8\%$ on GLM-5 at $\ret{2}$, and $7.9\%$ and $5.1\%$ at $\ret{4}$; and at retention $\ret{r}$ it is faster than IndexCache at $\ret{2r}$ while attending twice as many fresh selections, so a halving of the indexer budget comes at no speed cost. At $128$K it reaches $1.32\times$ the unmodified model's throughput at $\ret{4}$, above IndexCache's $1.28\times$ at $\ret{8}$.
  \item \textbf{A shared-primitive fused kernel, and the case for co-design.} On DeepSeek-V3.2 a fused decode kernel runs the sparse attention and the indexer's prefix scan of a shared step in one launch, a per-request win (\Cref{sec:res-fused}); it does not help GLM-5, which has fewer index heads per rank. Because the indexer scores the same group bilinear interaction as attention, this is evidence that DSA models should be co-designed around that shared primitive: a direction we argue for and partly realize, not a model we train.
  \item \textbf{The quality cost of staleness is small and recoverable.} Across five long-context suites (\Cref{sec:res-cost}), attending an older selection stays within a few points of IndexCache at every retention it reports; the arm with twice the fresh selections, now the faster of the two, matches IndexCache on LongBench and LongBench~v2 and is within $3$ points on RULER (\Cref{sec:res-frontier}), and distilling only the indexers, applied identically to both methods, recovers little at the mild retentions IndexCache reports but rescues the collapse at aggressive retention (\Cref{sec:res-recover}).
\end{itemize}

\section{Background}
\label{sec:background}

\paragraph{A DSA layer at decode time.}
Let $N$ be the number of layers and $\mathbf{h}_\ell$ the input of layer $\ell$ for the current token. With $L$ tokens in the cache, layer $\ell$ performs three steps in order: (i) the indexer scores every cached token with \Cref{eq:indexer}, reading an index-key cache of $L$ entries; (ii) a top-$k$ over the $L$ scores produces the selection $\Isel_\ell$; (iii) sparse Multi-head Latent Attention runs over the $k$ selected latent KV rows. Only step (i) grows with $L$.

\paragraph{IndexCache.}
IndexCache~\citep{bai2026indexcache} splits the layers into \emph{full} layers, which run step (i), and \emph{shared} layers, which copy the selection of the nearest preceding full layer. A uniform pattern with period $r$ keeps a fraction $\ret{r}$ of the indexers; a greedy search against a calibration loss does better at equal retention, and the paper publishes searched patterns for GLM-5. GLM-5.2 adopts the uniform pattern with $r{=}4$ as IndexShare~\citep{zai2026glm52}. A full layer still waits for its own scan. IndexCache also describes a training-aware variant that distills each retained indexer toward the selections of the layers it serves.

\paragraph{Why a neighbouring layer's selection is usable.}
The dependency between the indexer and attention is on the \emph{support} of the softmax, not on any computation they share, and it disappears the moment attention takes its support from a selection that already exists when $\mathbf{h}_\ell$ does. Such a selection is available if adjacent layers select similar tokens. On real activations (GovReport, $k{=}2048$) the Jaccard overlap between consecutive layers' selections stays near $0.42$ on DeepSeek-V3.2 and $0.26$ on GLM-5 from $16$K to $128$K; whether that suffices is measured directly in \Cref{sec:res-cost}. IndexCache uses this by copying a selection forward; \MLI{} attends the previous full layer's selection while the current one is computed. Only the indexer's \emph{placement} moves, not its arithmetic.

\section{\methodname}
\label{sec:method}

\subsection{Schedule}
\label{sec:method-schedule}
We describe which selection each layer attends. Let $\Isel_f$ denote the top-$2048$ tokens the indexer of full layer $f$ selects. Let $F\subseteq\{1,\dots,N\}$ be the set of full layers, those that run the indexer; standard DSA has $F=\{1,\dots,N\}$, and IndexCache with retention $\ret{r}$ has $|F|=N/r$. For any layer $\ell$ let $p(\ell)$ be the nearest full layer at or before $\ell$, so that $p(\ell)=\ell$ when $\ell\in F$. Let $\pi_k$ step back $k$ full layers along $F$ (the \emph{lag depth}), so $\pi_0$ is the identity. For a full layer $f$ and a depth $k\ge0$ let $\pi_k(f)$ be the full layer $k$ positions before $f$ in $F$, with $\pi_0(f)=f$. Every layer attends the selection of one full layer:
\begin{equation}
  \text{layer } \ell \text{ attends } \Isel_{\sigma(\ell)}, \qquad \sigma(\ell) \;=\; \pi_k\big(p(\ell)\big).
  \label{eq:schedule}
\end{equation}
The retention $r$ and the depth $k$ are independent. With $k{=}0$ every layer attends the selection of its own full layer: this is standard DSA when $r{=}1$ and IndexCache when $r{>}1$. With $k{\ge}1$ a full layer attends the selection made $k$ full layers earlier, and a shared layer attends what its full layer attended. This is \MLI{} when $r{=}1$ and the composition of the two methods when $r{>}1$: \MLI{} acts on the full layers, IndexCache's copies act on the shared layers. The first $k$ full layers of the forward pass have no predecessor in $F$ and run the standard schedule.

The point of \Cref{eq:schedule} with $k\ge1$ is independence. The indexer of full layer $f$ reads $\mathbf{h}_f$ and its output is first consumed by the layer that follows $f$ in $F$; nothing in layer $f$ waits for it, so the indexer can execute while the attention, the output projection and the feed-forward block of layer $f$ execute (\Cref{fig:schedule} in the appendix draws the layer schedule). Two properties separate this from IndexCache: under \MLI{} a selection is at most $k$ full layers old and every full layer attends a selection recomputed for the current token, whereas under IndexCache a shared layer attends a selection up to $r{-}1$ layers old that was never recomputed for it. One serial scan per decode step remains, the first full layer of the forward pass; in a model whose first layers use dense attention it could be issued alongside the last dense layer, but DeepSeek-V3.2 and GLM-5 carry an indexer at every layer, so we keep it serial.

\subsection{Realization inside the decode graph}
\label{sec:method-realization}
\begin{figure}[t]
\centering
\resizebox{\linewidth}{!}{%
\begin{tikzpicture}[x=1cm,y=1cm,font=\scriptsize,
  main/.style={draw=blue!50!black,fill=blue!8,rounded corners=1pt,minimum height=6mm,inner sep=2pt,align=center,anchor=west},
  side/.style={draw=red!55!black,fill=red!10,rounded corners=1pt,minimum height=6mm,inner sep=2pt,align=center,anchor=west},
  lbl/.style={font=\scriptsize\bfseries,anchor=east},
  note/.style={font=\tiny,red!55!black},
  edge/.style={-{Stealth[length=1.8mm]},line width=0.7pt,red!55!black,rounded corners=2pt}]
\node[lbl] at (-0.15,1.2) {main stream};
\node[lbl] at (-0.15,0.0) {side stream};
\node[main,minimum width=1.8cm] (a1) at (0.0,1.2) {attention$_\ell$\\reads $\mathcal{S}_{\ell-1}$};
\node[main,minimum width=1.0cm] (o1) at (1.9,1.2) {$W_o$};
\node[main,minimum width=2.2cm] (m1) at (3.0,1.2) {MoE$_\ell$};
\node[side,minimum width=4.45cm] (i1) at (0.15,0.0) {indexer$_\ell$: scan $L$ keys, top-$k$ $\to\mathcal{S}_\ell$};
\node[main,minimum width=1.8cm] (a2) at (5.6,1.2) {attention$_{\ell+1}$\\reads $\mathcal{S}_{\ell}$};
\node[main,minimum width=1.0cm] (o2) at (7.5,1.2) {$W_o$};
\node[main,minimum width=2.2cm] (m2) at (8.6,1.2) {MoE$_{\ell+1}$};
\node[side,minimum width=4.45cm] (i2) at (5.75,0.0) {indexer$_{\ell+1}$: scan, top-$k$ $\to\mathcal{S}_{\ell+1}$};
\draw[edge] (0.3,0.9) -- (0.3,0.3);
\node[note,anchor=north west] at (0.05,-0.35) {fork: $\mathbf{h}_\ell$ ready};
\draw[edge] (i1.east) -- (5.4,0.0) -- (5.4,1.2) -- (a2.west);
\node[note,anchor=north east] at (5.35,-0.35) {join: $\mathcal{S}_\ell$ consumed by layer $\ell{+}1$};
\draw[edge] (5.9,0.9) -- (5.9,0.3);
\draw[gray] (0.0,1.75) -- (5.2,1.75); \node[gray,anchor=south] at (2.6,1.78) {layer $\ell$};
\draw[gray] (5.6,1.75) -- (10.8,1.75); \node[gray,anchor=south] at (8.2,1.78) {layer $\ell{+}1$};
\draw[-{Stealth[length=1.5mm]},gray] (0.0,-0.85) -- (10.8,-0.85); \node[gray,anchor=north east,font=\tiny] at (10.8,-0.88) {time};
\end{tikzpicture}}
\caption{Two-stream realization, depth $1$, one decode step across two full layers. The indexer of layer $\ell$ forks onto the side stream once $\mathbf{h}_\ell$ exists and runs under the attention, output projection and feed-forward block of layer $\ell$; its selection is joined at the prepare step of layer $\ell{+}1$, which attends it. The whole pattern is captured once into the decode CUDA graph.}
\label{fig:streams}
\end{figure}
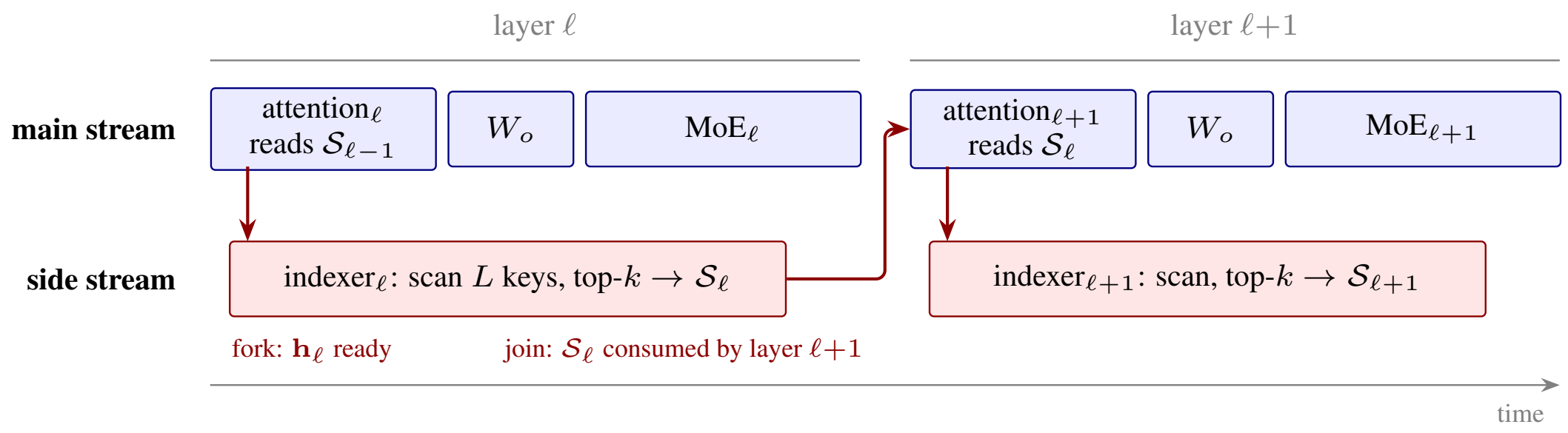

We implement \Cref{eq:schedule} in SGLang's DSA decode path; its portable realization changes no kernel (a fused kernel, faster on DeepSeek-V3.2, is \Cref{sec:res-fused}), and \Cref{fig:streams} shows one decode step. At a full layer, as soon as the layer input exists, the indexer (its projections, the scan over the paged index-key cache, and the top-$k$) is enqueued on a second CUDA stream while attention on the main stream proceeds with the selection that \Cref{eq:schedule} assigns to the layer. The second stream is joined at the layer that first consumes the new selection, $k$ full layers later, and references to the indexer's inputs and output are held until that join so the allocator cannot reuse their memory. The cross-stream edges are ordinary graph dependencies, so the whole schedule is captured once into the decode CUDA graph and replayed without host involvement; a per-layer registry carries selections between full layers, and shared layers receive the carried selection exactly as in IndexCache. Because the scan and the top-$k$ are the production kernels on the production buffers, each full layer produces the selection the standard schedule produces; only the layer that consumes it changes.

\subsection{Recovering selection quality by distillation}
\label{sec:method-distill}
\MLI{} changes exactly one thing, the selection a layer attends, and the cost of that change can be paid back without touching the backbone. We train only the indexers of the full layers, about $0.5\%$ of the parameters, so that the selection a full layer produces matches what the layers that will attend it would have selected for themselves: under \MLI{} those consumers are the layers from the next full layer up to the one after it, under IndexCache the full layer itself and its shared span. The teacher is the unmodified model's indexer at each consumer, frozen, read from the same forward pass. With $\mathcal{C}(f)$ the consumers of full layer $f$ and $p^{c},\hat p^{f}$ the softmax teacher and student scores at consumer $c$, the per-layer loss is
\begin{equation}
  \mathcal{L}_f \;=\; \frac{1}{|\mathcal{C}(f)|}\sum_{c\in\mathcal{C}(f)} \Big[\, \lambda\,\mathrm{KL}\!\big(p^{c}\,\|\,\hat p^{f}\big) \;+\; (1-\lambda)\,\mathcal{L}_{\mathrm{PL}}\!\big(\pi^{c};\,\hat I^{f}\big) \Big],
  \label{eq:distill-main}
\end{equation}
where $\mathcal{L}_{\mathrm{PL}}$ is a listwise Plackett--Luce (ListMLE) term on the teacher's top-$k$ ranking $\pi^{c}$ and $\lambda{=}0.5$; it fixes the ranking where the top-$k$ is decided (\Cref{app:distill}). The same data, steps and loss are applied to IndexCache on its own consumer set, which is its training-aware variant, so distilled arms are always compared in pairs.

\section{Experimental Setup}
\label{sec:setup}

\paragraph{Models and serving.}
We serve DeepSeek-V3.2 ($61$ layers, $128$ attention heads, $64$ indexer heads) and GLM-5 ($78$ layers, $64$ attention heads, $32$ indexer heads) from their fp8 checkpoints with SGLang v0.5.16 on one node of eight H200 GPUs, with $8$-way tensor parallelism, an fp8 KV cache, CUDA graphs enabled for every arm and $k{=}2048$ (a data-parallel-attention replication is in \Cref{app:dp}).

\paragraph{Regimes and contexts.}
We report (i) per-request decode throughput at a single concurrent request; (ii) full-KV-cache decode throughput, where the request set is sized to the server's KV capacity and every prompt is first warmed into the prefix cache so that the measured phase is saturated decode; and (iii) prefill latency. Contexts are $10$K, $60$K and $120$K tokens on both models, plus $128$K on DeepSeek-V3.2 and $195$K ($199{,}680$ tokens) on GLM-5. Retention is uniform with $r\in\{2,4,8,16,32\}$ (IndexCache reports up to $\ret{8}$). On GLM-5 we also run IndexCache's published searched patterns ($40$ and $22$ full layers of $78$), with \MLI{} applied to the same patterns.

\paragraph{Protocol and fairness.}
Every arm of a cell (one model, one regime, one context) runs in the same job on the same node, differing only in model-override flags, so a speed difference reflects the mechanism and not the setup. Three controls guard the comparison. \emph{(i)~Drift:} the unmodified model is booted first and again last, and the cell is reported only if the two agree within $2\%$ (observed within $0.4\%$). \emph{(ii)~Engagement:} every \MLI{} arm must log, on every tensor-parallel rank, the marker that the overlapped path was captured into the decode graph, with no fallback marker (\Cref{app:gates}). \emph{(iii)~IndexCache parity:} IndexCache runs through its own released layer-skip knobs, and as a control its arms are also booted on the unmodified SGLang release in the same job, where our build matches the release within $0.2\%$ on the base model and is $0.4$ to $1.1\%$ \emph{slower} on the IndexCache arms (\Cref{app:parity}), so the tables understate IndexCache by at most $1.1$ percentage points. Distilled arms are always paired with IndexCache distilled by the same recipe on its own spans (\Cref{sec:method-distill}). Measurement noise is in \Cref{app:noise}, and paired bootstrap intervals for every quality contrast in \Cref{app:quality-ci}.

\paragraph{Quality.}
All evaluations are served through the same endpoints, with greedy decoding, and every delta is against the unmodified model on the same job. \emph{LongBench}~\citep{bai2024longbench}: eleven English tasks, $100$ examples each, inputs middle-truncated to $250{,}000$ characters, task-native metrics, macro-averaged. \emph{LongBench~v2}~\citep{bai2025longbench}: multiple-choice accuracy on the full $503$-question set, inputs middle-truncated to $120$K tokens, the official prompt, answer budget and answer extraction, each model's non-thinking chat format (\Cref{app:quality-full}). \emph{RULER}~\citep{hsieh2024ruler}: needle retrieval, variable tracking and common-word extraction at $4$K to $128$K tokens, the benchmark's own metrics. \emph{Needle retrieval}: ten numeric needles in haystacks of $32$K to $520$K characters. \emph{Decode perplexity}: perplexity of the model's own greedy $512$-token continuation after $4$K to $64$K tokens of GovReport text, five seeds; a deviation in either direction is drift.

\section{Results}
\label{sec:results}

\subsection{One free halving of retention}
\label{sec:res-speed}
Our main result is per-request decode throughput, IndexCache and IndexCache$+$\MLI{} side by side at every retention and context (\Cref{tab:decode-per-request}), at a single concurrent request. Each model uses its best realization of \MLI{} (\Cref{sec:res-fused}): the fused decode kernel on DeepSeek-V3.2 and the deferred-stream schedule on GLM-5, where the fused kernel does not help; both attend exactly the selections the serial schedule would. We read the table two ways. Across a row, the composition is faster than IndexCache at equal retention. The gain is largest where IndexCache keeps the most indexers: at $\ret{2}$ the composition decodes up to $14.5\%$ faster than IndexCache on DeepSeek-V3.2 and $9.8\%$ on GLM-5, at $\ret{4}$ up to $7.9\%$ and $5.1\%$, and it narrows toward parity at $\ret{32}$ as fewer scans remain to hide. It grows with context and holds on IndexCache's searched patterns. Down a column, IndexCache$+$\MLI{} at $\ret{r}$ is \emph{faster} than IndexCache at $\ret{2r}$: at $128$K and $\ret{4}$ it is $3\%$ faster than IndexCache at $\ret{8}$ ($1.32\times$ the unmodified model against IndexCache's $1.28\times$) despite half the retained indexers, and the same holds at $\ret{2}$ and $\ret{8}$ (on GLM-5 by $2$ to $3\%$). Which mechanism buys the halving differs by model: deferred overlap on GLM-5, the fused kernel on DeepSeek-V3.2 (\Cref{sec:res-fused}). Prefill latency, which \MLI{} leaves unchanged, is in \Cref{app:ttft}.

\begin{table}[t]
\centering\small
\setlength{\tabcolsep}{4pt}
\renewcommand{\arraystretch}{0.9}
\begin{tabular}{lrrrrrrrr}
\toprule
\textbf{DeepSeek-V3.2} & \multicolumn{2}{c}{10K} & \multicolumn{2}{c}{60K} & \multicolumn{2}{c}{120K} & \multicolumn{2}{c}{128K} \\
retention & \IC{} & +\MLI{} & \IC{} & +\MLI{} & \IC{} & +\MLI{} & \IC{} & +\MLI{} \\
\midrule
none (base, tok/s) & \multicolumn{2}{c}{89.1} & \multicolumn{2}{c}{88.0} & \multicolumn{2}{c}{85.7} & \multicolumn{2}{c}{85.5} \\
all (\MLI{} alone) & \multicolumn{2}{c}{+21.5} & \multicolumn{2}{c}{+19.8} & \multicolumn{2}{c}{+26.1} & \multicolumn{2}{c}{+25.1} \\
\midrule
$\ret{2}$ & +11.5 & +25.3 & +12.3 & +25.2 & +13.6 & +30.1 & +13.7 & +29.8 \\
$\ret{4}$ & +19.1 & +27.3 & +20.5 & +28.3 & +22.8 & +32.5 & +22.7 & +32.3 \\
$\ret{8}$ & +23.4 & +28.5 & +25.2 & +29.8 & +27.9 & +33.7 & +28.0 & +33.7 \\
$\ret{16}$ & +25.3 & +26.3 & +27.1 & +28.2 & +30.2 & +31.6 & +30.2 & +31.7 \\
$\ret{32}$ & +26.1 & +26.9 & +28.2 & +28.5 & +31.4 & +32.2 & +31.4 & +32.2 \\
\bottomrule
\end{tabular}
\vspace{4pt}
\begin{tabular}{lrrrrrrrr}
\toprule
\textbf{GLM-5} & \multicolumn{2}{c}{10K} & \multicolumn{2}{c}{60K} & \multicolumn{2}{c}{120K} & \multicolumn{2}{c}{195K} \\
retention & \IC{} & +\MLI{} & \IC{} & +\MLI{} & \IC{} & +\MLI{} & \IC{} & +\MLI{} \\
\midrule
none (base, tok/s) & \multicolumn{2}{c}{68.5} & \multicolumn{2}{c}{67.8} & \multicolumn{2}{c}{66.5} & \multicolumn{2}{c}{65.0} \\
all (\MLI{} alone) & \multicolumn{2}{c}{+14.6} & \multicolumn{2}{c}{+14.4} & \multicolumn{2}{c}{+15.6} & \multicolumn{2}{c}{+16.8} \\
\midrule
$\ret{2}$ & +8.2 & +17.6 & +8.7 & +17.9 & +9.8 & +19.8 & +10.6 & +21.4 \\
$\ret{4}$ & +13.7 & +18.8 & +14.2 & +19.2 & +15.9 & +21.6 & +17.7 & +23.7 \\
$\ret{8}$ & +16.3 & +19.4 & +17.1 & +20.4 & +19.3 & +22.8 & +21.3 & +25.0 \\
$\ret{16}$ & +18.1 & +20.0 & +18.9 & +20.9 & +21.2 & +23.3 & +23.4 & +25.7 \\
$\ret{32}$ & +18.8 & +20.2 & +19.7 & +21.1 & +22.1 & +23.5 & +24.4 & +25.8 \\
\midrule
$\ret{2}$ searched & +8.4 & +17.2 & +8.7 & +17.5 & +9.8 & +19.3 & +10.9 & +21.1 \\
$\ret{4}$ searched & +13.4 & +18.5 & +13.9 & +19.1 & +15.5 & +21.4 & +17.2 & +23.4 \\
\bottomrule
\end{tabular}
\vspace{4pt}
\caption{Per-request decode throughput at single concurrency. Base row: tokens/s of the unmodified model; all other cells: percentage change over base in the same cell, positive = faster, for \IC{} alone and for \IC{}$+$\MLI{} at the same retention. All arms of a cell run in one job on one node. $\ret{16}$ and $\ret{32}$ extend \IC{}'s published sweep ($\ret{2}$--$\ret{8}$). DeepSeek-V3.2 $+$\MLI{} cells use the fused-kernel realization (\Cref{sec:res-fused}); GLM-5 uses the deferred-stream realization, which is faster there; both are selection-identical to the production top-$k$.}
\label{tab:decode-per-request}
\end{table}

\paragraph{Full KV cache.}
\Cref{tab:decode-saturation} (appendix) repeats the sweep with the KV cache full; the gain from \MLI{} tracks the number of requests decoding together, largest where the context is long enough to hold few (\Cref{app:saturation}).

\subsection{What limits the gain}
\label{sec:res-mechanism}
With every indexer retained ($r{=}1$) and one request, \MLI{} raises decode throughput by $19.8$ to $26.1\%$ on DeepSeek-V3.2 (the fused kernel) and $14.4$ to $16.8\%$ on GLM-5 (the deferred stream) across $10$K to $195$K (\Cref{tab:decode-per-request}, second row). The ceiling, obtained by removing every indexer that can be removed (IndexCache at $\ret{32}$ retains two), is $+31\%$ on DeepSeek-V3.2 and $+24\%$ on GLM-5 at the longest context, so \MLI{} hides about four-fifths of the exposed scan on DeepSeek-V3.2 and about two-thirds on GLM-5. The first full layer's scan stays serial (no earlier layer to overlap), so even the ceiling is not fully reachable. Two ablations (\Cref{tab:mechanism}) rule out scheduling as the further limit: a high-priority CUDA stream and a deeper $k{=}2$ or $3$ window both change nothing. What remains is memory bandwidth: the scan hides only in the HBM bandwidth attention and the feed-forward block leave idle, which is why the gain vanishes once requests saturate it (\Cref{tab:decode-saturation}, $64$ requests; not profiled directly).

\subsection{A fused kernel where bandwidth allows}
\label{sec:res-fused}
\MLI{} has two realizations, both attending the production top-$k$. The \emph{deferred stream} changes no kernel and issues the next layer's indexer on a second CUDA stream; the \emph{fused kernel} runs the sparse attention and the prefix scan of the shared step in one launch, so the scan never competes for SMs with a separate launch. On DeepSeek-V3.2 the fused kernel is faster at every retention and context, by $4$ to $6$ percentage points (composition at $\ret{4}$, $128$K: $+32.3\%$ against $+27.5\%$), and \Cref{tab:decode-per-request} uses it there. On GLM-5 it loses $2$ to $4$ percentage points to the deferred stream, which \Cref{tab:decode-per-request} uses instead: GLM-5 has half as many index heads per rank and a smaller per-layer feed-forward window, so the single launch cannot hide the scan. Selections are bit-identical, so quality (\Cref{sec:res-cost}) is unchanged. The fused kernel is per-request only: at full-KV-cache batch sizes it is issue-bound (about $20\%$ slower than the stream at $64$ requests), so \Cref{tab:decode-saturation} uses the deferred stream.

\subsection{What \MLI{} costs}
\label{sec:res-cost}
\begin{table}[t]
\centering\small
\setlength{\tabcolsep}{4pt}
\renewcommand{\arraystretch}{0.9}
\begin{tabular}{lrrrrrr}
\toprule
\textbf{DeepSeek-V3.2} & \multicolumn{2}{c}{LongBench} & \multicolumn{2}{c}{RULER} & \multicolumn{2}{c}{LongBench v2} \\
retention & \IC{} & +\MLI{} & \IC{} & +\MLI{} & \IC{} & +\MLI{} \\
\midrule
none (base, score) & \multicolumn{2}{c}{0.449} & \multicolumn{2}{c}{0.975} & \multicolumn{2}{c}{0.501} \\
all (\MLI{} alone) & \multicolumn{2}{c}{-0.0} & \multicolumn{2}{c}{-2.1} & \multicolumn{2}{c}{-0.4} \\
all (\MLI{} alone), distilled & \multicolumn{2}{c}{-0.4} & \multicolumn{2}{c}{-4.7} & \multicolumn{2}{c}{-4.4} \\
\midrule
$\ret{2}$ & \textbf{-0.2} & -0.3 & \textbf{-2.7} & -4.0 & \textbf{+0.8} & +0.4 \\
$\ret{2}$ distilled & \textbf{-0.0} & -0.8 & \textbf{-1.1} & -3.7 & -0.8 & \textbf{+0.0} \\
$\ret{4}$ & -0.3 & \textbf{+0.5} & -0.9 & \textbf{-0.4} & -1.6 & \textbf{-0.4} \\
$\ret{4}$ distilled & \textbf{+0.2} & -0.4 & \textbf{-2.0} & -4.0 & -0.8 & -0.8 \\
$\ret{8}$ & \textbf{-0.1} & -1.1 & -0.7 & \textbf{+0.3} & -0.4 & \textbf{+0.4} \\
$\ret{8}$ distilled & -0.6 & \textbf{-0.5} & \textbf{-0.9} & -2.7 & \textbf{+0.4} & -0.4 \\
\bottomrule
\end{tabular}
\vspace{2pt}
\begin{tabular}{lrrrrrr}
\toprule
\textbf{GLM-5} & \multicolumn{2}{c}{LongBench} & \multicolumn{2}{c}{RULER} & \multicolumn{2}{c}{LongBench v2} \\
retention & \IC{} & +\MLI{} & \IC{} & +\MLI{} & \IC{} & +\MLI{} \\
\midrule
none (base, score) & \multicolumn{2}{c}{0.453} & \multicolumn{2}{c}{0.941} & \multicolumn{2}{c}{0.477} \\
all (\MLI{} alone) & \multicolumn{2}{c}{-0.2} & \multicolumn{2}{c}{+2.3} & \multicolumn{2}{c}{-0.8} \\
all (\MLI{} alone), distilled & \multicolumn{2}{c}{-1.3} & \multicolumn{2}{c}{+2.9} & \multicolumn{2}{c}{+0.8} \\
\midrule
$\ret{2}$ & -1.2 & \textbf{-1.1} & \textbf{+2.9} & +2.1 & \textbf{+1.2} & +0.4 \\
$\ret{2}$ distilled & -0.2 & \textbf{+0.4} & +2.0 & \textbf{+2.2} & \textbf{+2.9} & -2.1 \\
$\ret{4}$ & \textbf{-0.2} & -0.4 & +0.7 & \textbf{+1.6} & \textbf{+2.5} & -0.4 \\
$\ret{4}$ distilled & -0.8 & \textbf{+0.9} & \textbf{+0.9} & -1.1 & -0.8 & \textbf{+2.1} \\
$\ret{8}$ & \textbf{-0.0} & -1.4 & \textbf{+1.6} & -1.3 & \textbf{+0.8} & -1.7 \\
$\ret{8}$ distilled & -0.4 & \textbf{+0.4} & +1.2 & \textbf{+1.2} & \textbf{+0.0} & -2.1 \\
\bottomrule
\end{tabular}
\vspace{2pt}
\caption{Quality at the retentions \IC{} reports, \IC{} and \IC{}$+$\MLI{} side by side. Base row: absolute score of the unmodified model; other cells: percentage change over base, positive = better, with the better arm of each \IC{}/$+$\MLI{} pair in bold. LongBench: macro-F1 over 11 tasks; RULER: mean over tasks and lengths 4K--128K; LongBench~v2: accuracy on the full 503-question set (one question is $0.2$ points, so differences under a point are within its resolution). Distilled rows: indexer-only distillation, same recipe and budget for both methods. Needle retrieval is $1.00$ for every arm shown (it collapses only at $\ret{16}$--$\ret{32}$, outside this table; see \Cref{tab:quality}) and decode perplexity is within $\pm 6\%$ ($\pm 11\%$ on GLM-5) of base; \Cref{tab:quality} has every arm, metric and retention.}
\label{tab:quality-main}
\end{table}

The only behavioral change is which selection a layer attends. \Cref{tab:quality-main} pairs the two methods at the retentions IndexCache reports, under the serving protocol where every retained layer computes its selection at prefill and copies only at decode (\Cref{app:quality-full}). With every indexer retained, \MLI{} alone is within $0.4$--$0.8$ points of base on LongBench and LongBench~v2, and on RULER $2.1$ below on DeepSeek-V3.2 / $2.3$ above on GLM-5, inside the models' own base spread. Retention is cheap at $\ret{2}$ and $\ret{4}$: both arms stay within about $3$ points of base on every suite. On top of IndexCache, \MLI{} costs at most a point or two (\Cref{tab:quality-main}); the paired $95\%$ intervals resolve differences that small only on RULER, spanning zero on LongBench and LongBench~v2 (\Cref{app:quality-ci}).

\subsection{Recovering it}
\label{sec:res-recover}
Distillation earns its place unevenly across the retention range, and where it does is exactly where the free halving matters most. At the mild retentions IndexCache reports the cost is already small, so distilling only the indexers (the recipe of \Cref{sec:method-distill}, applied identically to both methods) has correspondingly little to recover and is neutral to slightly negative: distilled and undistilled arms are within $2.5$ points of each other on LongBench and LongBench~v2, and on RULER the distilled composition on DeepSeek-V3.2 sits $3$ to $4$ points below the unmodified model against $1$ to $2$ for distilled IndexCache (the intervals in \Cref{app:quality-ci} exclude zero for the former and not the latter). At the cliff it is decisive: at $\ret{16}$ on GLM-5, distilling the indexers turns IndexCache's collapse (LongBench $-11.5$, RULER $-61$, needle $0.00$) into a recovery (LongBench $-0.1$, RULER $-7.4$, needle $1.00$; \Cref{tab:quality}), moving the usable range one halving further. Distillation swaps only indexer weights, so decode speed is unchanged. One mild-retention cell moves more (GLM-5 at $\ret{2}$ on LongBench~v2, $-2.5$; a single-seed swing).

\subsection{At equal speed}
\label{sec:res-frontier}
The halving pair no longer runs at equal speed (the composition at $\ret{r}$ is the faster arm; \Cref{tab:decode-per-request}), and it gives up nothing in quality to buy that speed: both arms sit within a few points of base, and every $95\%$ interval spans zero (\Cref{app:quality-ci}, \Cref{tab:equal-speed}) except one RULER pair on DeepSeek-V3.2 ($-3.1$, $[-5.9,-0.6]$). So within the retentions IndexCache reports the extra freshness comes free; one halving beyond, at the cliff (\Cref{sec:res-recover}), it also avoids collapse, as IndexCache degenerates at $\ret{16}$ on GLM-5 and $\ret{32}$ on DeepSeek-V3.2 while the composition one halving above does not (\Cref{tab:quality}). These are single runs.

\section{Related Work}
\label{sec:related}

\paragraph{Sparse attention and selection.}
Fixed-pattern sparse attention~\citep{child2019generating,beltagy2020longformer,zaheer2020big} attends a static subset; the learned variants adapt it: NSA~\citep{yuan2025native}, MoBA~\citep{lu2026moba}, InfLLM-V2~\citep{zhao2026infllm} and SeerAttention~\citep{gao2024seerattention} learn to select blocks or tokens, DSA~\citep{liu2025deepseek} with a per-layer lightning indexer; Quest~\citep{tang2024quest} estimates importance from cached key statistics and MInference~\citep{jiang2024minference} from offline-assigned per-head sparse patterns; eviction and offloading~\citep{zhang2023h2o,xiao2024efficient,li2024snapkv,zhu2025oraclekv,luo2025headinfer} shrink or relocate the cache rather than the read; and \citet{sadhukhan2026kinetics} argue that at test time sparse attention sets the cost per token. All of these run selection before attention. Distillation into sparse attention has targeted whole attention layers~\citep{sundarapandiyan2025foreign}; we distill only the indexers of a model that already is sparse, toward its own selections. IndexCache~\citep{bai2026indexcache} shares one DSA layer's selection with neighbouring layers; TidalDecode~\citep{yang2025tidaldecode} and Kascade~\citep{deshmukh2025kascade} reuse attention-derived selections across layers in models without an indexer. These reduce how often selection runs; \MLI{} reduces how long a layer waits for it, and composes with them.

\paragraph{Selection and attention as one primitive.}
Attention's logit and the indexer's score are both low-rank bilinear interactions of the layer input, so the indexer is a second consumer of what attention already computes. Selective Attention~\citep{leviathan2025selective} makes a related point at training time: reusing attention's own scores as the selection signal rather than a separate module. \MLI{} exploits only the cross-layer overlap of the resulting selections, but our serving results argue for the tighter co-design: on DeepSeek-V3.2 a single fused kernel computes the attention and the indexer scan in one launch. Training a model whose indexer and attention share that primitive from the start is the natural next step.

\section{Limitations}
\label{sec:limitations}
The gain of \MLI{} falls with the number of requests decoding together. The fused kernel helps only DeepSeek-V3.2; on GLM-5 it loses to the deferred stream (\Cref{sec:res-fused}). We evaluate GLM-5 as the family's representative: GLM-5.1 shares its base model and DSA architecture (only the checkpoint differs), so our results carry over; GLM-5.2 ships IndexCache natively as IndexShare, so a comparison there would not be on equal footing and we leave it out of scope.

\section{Conclusion}
DSA places a lightning indexer in front of every attention layer, and the indexer scan is a serial dependency of the step it feeds. It need not be. Adjacent DSA layers select almost the same tokens, so a layer can attend the selection an earlier indexer already produced and run its own scan alongside attention rather than before it. The change is drop-in: no retraining, no new weights, and it composes with IndexCache. Serving DeepSeek-V3.2 and GLM-5, it makes IndexCache faster at every retention, by up to $14.5\%$ in decode throughput where the indexers are densest, and buys a further halving of the indexer budget for free, since at half the retention the composition is already faster than IndexCache while attending twice as many fresh selections. The indexer only recomputes the bilinear interaction attention already forms, so the scan is not a separate cost to hide but a second reader of one primitive; the step this invites is to co-design DSA around that primitive rather than place a selector ahead of it.

\bibliographystyle{plainnat}
\bibliography{refs}

\appendix

\section{Serving Protocol and Gates}
\label{app:gates}
\paragraph{Server.} SGLang v0.5.16 with our overlay, \texttt{--tp-size 8}, \texttt{--mem-fraction-static 0.83}, decode CUDA graphs captured for batch sizes $1$, $2$, $4$ (per-request cells) or $1$ to $64$ (full-KV-cache cells), context limit $163{,}840$ (DeepSeek-V3.2) or $202{,}752$ (GLM-5), one server boot per arm. Method arms differ from base only in JSON model-override arguments (\texttt{index\_topk\_freq}, \texttt{index\_topk\_pattern}, \texttt{index\_reuse\_lag\_refresh}, \texttt{index\_reuse\_lag\_refresh\_deferred}, \texttt{index\_reuse\_lag\_depth}, \texttt{index\_distilled\_weights}).

\paragraph{Regimes.} \emph{Per-request}: \texttt{sglang.benchmark.one\_batch\_server} at batch size $1$, input length $L$, $128$ output tokens; we report the last decode throughput and the last time-to-first-token. \emph{Full KV cache}: the request set is sized to the server's KV capacity, $n=\lfloor \text{max\_total\_tokens}/(L+1024)\rfloor$ capped at $64$; a first \texttt{bench\_serving} pass with the same prompts warms them into the radix cache, and a second pass decodes $1024$ tokens per request at concurrency $n$; we report the aggregate decode throughput of the second pass. (\texttt{bench\_serving}'s raw output throughput is prefill-dominated at $128$K and is not reported.)

\paragraph{Gates.} Every cell (one model, one regime, one context) runs all of its arms in one Slurm job on one node, in the order base, method arms, base again. A number is reportable only if the cell passes all of: (S) the result file exists and reports a positive steady throughput; (A) the closing base repeat is within $2\%$ of the opening one; (E) engagement: the server log contains the marker proving the intended path ran (\texttt{first 'lag\_refresh\_deferred'} for \MLI{}, \texttt{first 'lag\_depth\_k'} for depth $k$) and none of the fallback strings that the overlay emits when a path is bypassed; (I) plausibility: an arm that only removes or hides a scan cannot exceed the indexer-free ceiling and cannot be slower than base by more than a tolerance; (C) the overlay build's base matches the unmodified release on the same node (\Cref{app:parity}). Verdicts are written next to the cell; the table generator reads only PASS arms from cells whose two base runs pass (S), (E) and (A).

\paragraph{Selection identity.} The stream realization runs the production scan and top-$k$ kernels on the production buffers, so its selections are the serial schedule's by construction.

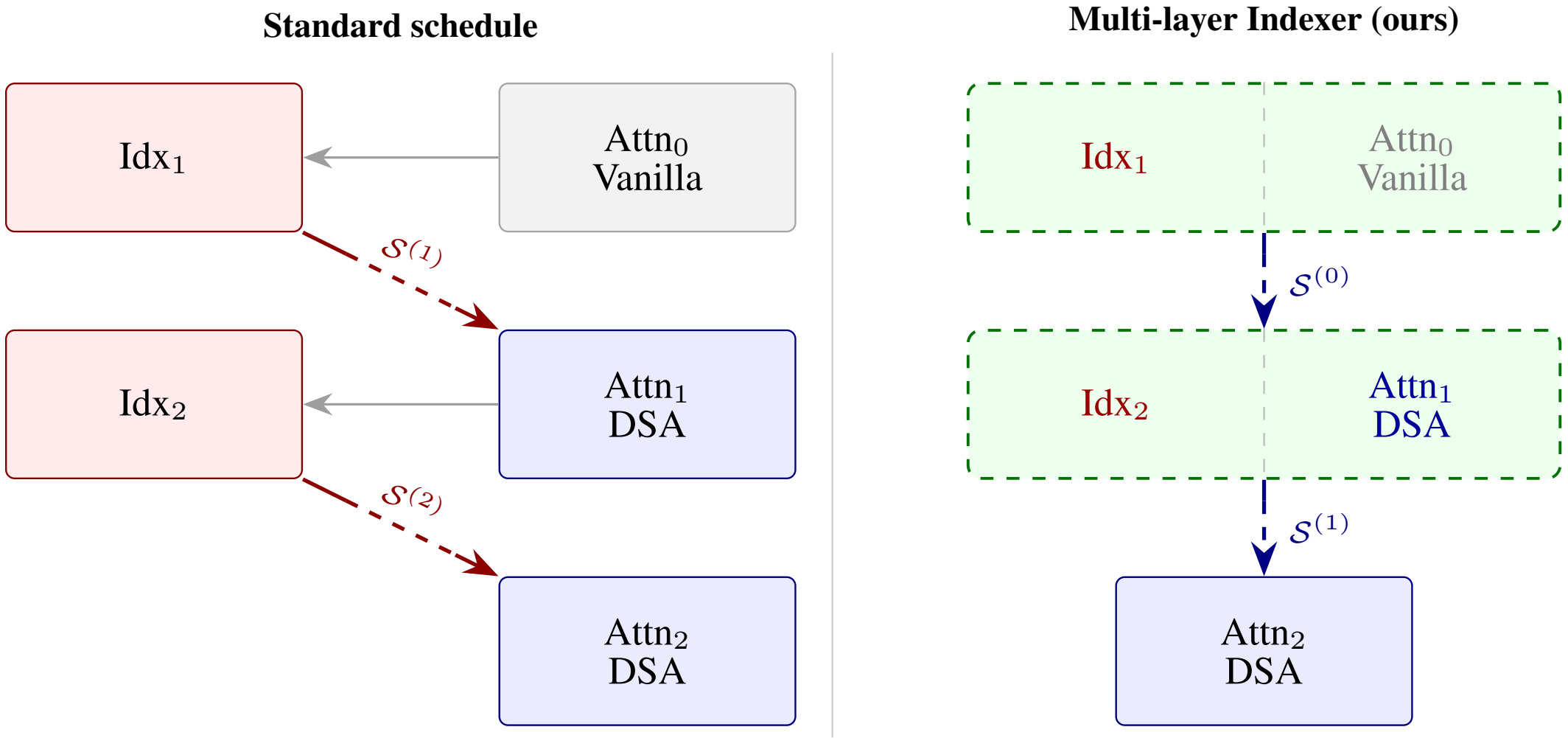
\begin{figure}[t]
\centering
\resizebox{\linewidth}{!}{%
\begin{tikzpicture}[
  ibox/.style={draw=red!50!black, rectangle, rounded corners=2pt,
               minimum width=2.4cm, minimum height=1.2cm,
               fill=red!8, align=center, font=\small},
  abox/.style={draw=blue!50!black, rectangle, rounded corners=2pt,
               minimum width=2.4cm, minimum height=1.2cm,
               fill=blue!8, align=center, font=\small},
  vbox/.style={draw=black!35, rectangle, rounded corners=2pt,
               minimum width=2.4cm, minimum height=1.2cm,
               fill=black!5, align=center, font=\small},
  harr/.style={-{Stealth[length=2.4mm]}, line width=0.55pt, black!38},
  tarr/.style={-{Stealth[length=2.8mm]}, line width=0.9pt,  red!55!black},
  larr/.style={-{Stealth[length=2.8mm]}, line width=0.9pt,  blue!50!black},
]

\node[font=\footnotesize\bfseries, anchor=south] at (2.0, 5.85) {Standard schedule};

\node[ibox] (ind1)  at (0.0, 5.0) {Idx$_1$};
\node[vbox] (attn0) at (4.0, 5.0) {Attn$_0$\\[-1pt]Vanilla};
\node[ibox] (ind2)  at (0.0, 3.0) {Idx$_2$};
\node[abox] (att1)  at (4.0, 3.0) {Attn$_1$\\[-1pt]DSA};
\node[abox] (att2)  at (4.0, 1.0) {Attn$_2$\\[-1pt]DSA};

\draw[harr] (attn0.west) -- (ind1.east);
\draw[harr] (att1.west)  -- (ind2.east);
\foreach \s/\e/\lbl in {
    ind1.south east/att1.north west/$\mathcal{S}^{(1)}$,
    ind2.south east/att2.north west/$\mathcal{S}^{(2)}$}{
  \draw[-, line width=0.9pt, red!55!black] (\s) -- ($(\s)!0.22!(\e)$);
  \draw[dashed, line width=0.9pt, red!55!black]
      ($(\s)!0.22!(\e)$)
      -- node[above, sloped, font=\scriptsize, pos=0.5]{\lbl}
      ($(\s)!0.78!(\e)$);
  \draw[-{Stealth[length=2.8mm]}, line width=0.9pt, red!55!black]
      ($(\s)!0.78!(\e)$) -- (\e);
}

\draw[black!18, thin] (5.5, 0.3) -- (5.5, 5.85);

\def\rx{9.0}
\node[font=\footnotesize\bfseries, anchor=south] at (\rx, 5.85) {\methodname{} (ours)};

\node[draw=green!45!black, dashed, rectangle, rounded corners=3pt,
      minimum width=4.8cm, minimum height=1.2cm, fill=green!7,
      line width=0.65pt] (top) at (\rx, 5.0) {};
\node[draw=green!45!black, dashed, rectangle, rounded corners=3pt,
      minimum width=4.8cm, minimum height=1.2cm, fill=green!7,
      line width=0.65pt] (mid) at (\rx, 3.0) {};
\node[abox] (bot) at (\rx, 1.0) {Attn$_2$\\[-1pt]DSA};

\node[font=\small, text=red!60!black,  align=center] at (\rx-1.2, 5.0) {Idx$_1$};
\node[font=\small, text=black!50,      align=center] at (\rx+1.2, 5.0) {Attn$_0$\\[-1pt]Vanilla};
\node[font=\small, text=red!60!black,  align=center] at (\rx-1.2, 3.0) {Idx$_2$};
\node[font=\small, text=blue!60!black, align=center] at (\rx+1.2, 3.0) {Attn$_1$\\[-1pt]DSA};

\draw[black!25, thin, dashed]
    (top.north -| top.center) -- (top.south -| top.center);
\draw[black!25, thin, dashed]
    (mid.north -| mid.center) -- (mid.south -| mid.center);

\coordinate (r1s) at (top.south -| top.center);
\coordinate (r1e) at (mid.north -| mid.center);
\coordinate (r2s) at (mid.south -| mid.center);
\coordinate (r2e) at (bot.north);
\foreach \s/\e/\lbl in {r1s/r1e/$\mathcal{S}^{(0)}$, r2s/r2e/$\mathcal{S}^{(1)}$}{
  \draw[-, line width=0.9pt, blue!50!black] (\s) -- ($(\s)!0.22!(\e)$);
  \draw[dashed, line width=0.9pt, blue!50!black]
      ($(\s)!0.22!(\e)$)
      -- node[right=2pt, font=\scriptsize]{\lbl}
      ($(\s)!0.78!(\e)$);
  \draw[-{Stealth[length=2.8mm]}, line width=0.9pt, blue!50!black]
      ($(\s)!0.78!(\e)$) -- (\e);
}

\end{tikzpicture}}
\caption{Layer schedule with depth $k{=}1$, drawn for a dense layer followed by two DSA layers. Left: standard DSA. Attention at layer $\ell$ starts only after the indexer of layer $\ell$ has produced $\Isel_\ell$. Right: \methodname{}. The indexer of layer $\ell{+}1$ and the attention of layer $\ell$ read the same hidden state and run concurrently (green step); a selection is attended one step after it is produced.}
\label{fig:schedule}
\end{figure}

\section{Ablations: Stream Priority and Lag Depth}
\label{app:ablations}
\begin{table}[h]\centering\small
\setlength{\tabcolsep}{4pt}
\caption{\MLI{} scheduling ablation with every indexer retained ($r{=}1$), the deferred-stream realization on the same overlay and recipe as \Cref{tab:decode-per-request} (\texttt{HH\_LAG\_NO\_ALT}). Per-request decode throughput relative to the unmodified model on the same node, drift-gated (base within $2\%$ of a closing base), at the two ends of the context range. Deepening the lag window to $k{=}2$ or $3$ full layers and giving the indexer a high-priority CUDA stream both leave the gain unchanged, so scheduling is not the limit. The DeepSeek cells here ran on nodes with a lower absolute base than \Cref{tab:decode-per-request} ($82.9$--$86.1$ vs $\sim85.5$ tok/s), so their percentages run a few percentage points high; the deferred arm reaches the same absolute throughput ($\sim100.7$ tok/s at $128$K, matching \Cref{sec:res-fused}).}
\label{tab:mechanism}
\begin{tabular}{lrrrr}
\toprule
 & \multicolumn{2}{c}{DeepSeek-V3.2} & \multicolumn{2}{c}{GLM-5} \\
 & 10K & 128K & 10K & 195K \\
\midrule
base (tok/s) & 86.1 & 82.9 & 69.2 & 65.8 \\
\MLI{} depth 1 & $+22.0$ & $+21.4$ & $+13.6$ & $+15.8$ \\
\MLI{} depth 1, high-priority stream & $+22.0$ & $+21.0$ & $+13.5$ & $+15.6$ \\
\MLI{} depth 2 & $+21.9$ & $+21.8$ & $+14.2$ & $+16.4$ \\
\MLI{} depth 3 & $+21.9$ & $+21.8$ & $+14.8$ & $+16.8$ \\
\bottomrule
\end{tabular}
\end{table}
Depth $2$ and $3$ track depth $1$ within about a percentage point at $r{=}1$ (above). We do not characterize quality for $k>1$: since deeper windows give no speed benefit, there is no reason to trade selection freshness for a larger lag. The indexer-free ceiling ($+31\%$ DeepSeek-V3.2, $+24\%$ GLM-5; \Cref{tab:decode-per-request}, $\ret{32}$) sits well above every row here, so what the scheduling ablations leave on the table is not scheduling.

\section{\IC{} Parity Control}
\label{app:parity}
\begin{table}[h]\centering\small
\caption{Unmodified SGLang release (stock) vs our build (ours), same job and node, per-request decode tok/s; \IC{} arms use the release's own \texttt{index\_topk\_freq}. Closing base sentinels: $0.0$ to $0.2\%$.}
\begin{tabular}{lrrrrrrrr}
\toprule
 & \multicolumn{4}{c}{DeepSeek-V3.2} & \multicolumn{4}{c}{GLM-5} \\
 & \multicolumn{2}{c}{10K} & \multicolumn{2}{c}{128K} & \multicolumn{2}{c}{10K} & \multicolumn{2}{c}{128K} \\
 & stock & ours & stock & ours & stock & ours & stock & ours \\
\midrule
base & 89.4 & 89.4 & 85.8 & 85.6 & 68.4 & 68.2 & 66.3 & 66.1 \\
\IC{} $\ret{2}$ & 100.6 & 99.6 & 98.2 & 97.4 & 74.8 & 74.0 & 73.5 & 72.7 \\
\IC{} $\ret{4}$ & 107.3 & 106.2 & 106.1 & 105.0 & 78.0 & 77.7 & 77.2 & 76.9 \\
\bottomrule
\end{tabular}
\end{table}
Our build is within $0.2\%$ of the release on base and $0.4$ to $1.1\%$ slower on the \IC{} arms. The \IC{} numbers in the main tables are from our build, so they understate stock \IC{} by up to $1.1$ percentage points; the equal-retention margins in \Cref{tab:decode-per-request} exceed that comfortably at $\ret{2}$ to $\ret{8}$ and shrink to within it at $\ret{16}$ and $\ret{32}$. The down-column result survives the correction: even against stock \IC{}, \IC{}+\MLI{} at $\ret{r}$ is faster than stock \IC{} at $\ret{2r}$, by $2$ to $6$ percentage points on DeepSeek-V3.2 and by up to $3$ on GLM-5.

\section{Full-KV-Cache Throughput}
\label{app:saturation}
\Cref{tab:decode-saturation} (appendix) repeats the sweep with the KV cache full. The number of requests decoding together is set by the context ($64$ at $10$K down to $6$ at $128$K on DeepSeek-V3.2, $43$ down to $2$ on GLM-5), and the \MLI{} gain follows it. At $64$ requests \MLI{} alone is $2.2\%$ slower than the unmodified model and the composition trails IndexCache by up to $4$ points; at every lower concurrency ($43$ and below) \MLI{} alone is positive ($4.3$ to $11.3\%$), the composition is within about $3$ points of IndexCache either way, and at the longest context it runs up to $6$ points above IndexCache at $\ret{2}$. The single-request gain therefore carries to the saturated server wherever the context holds few requests, the regime where the scan dominates the step; only when many short requests pack together does the feed-forward block saturate bandwidth by itself and the gain vanish (\Cref{sec:res-mechanism}).
\begin{table}[t]
\centering\small
\setlength{\tabcolsep}{4pt}
\renewcommand{\arraystretch}{0.9}
\begin{tabular}{lrrrrrrrr}
\toprule
\textbf{DeepSeek-V3.2} & \multicolumn{2}{c}{10K} & \multicolumn{2}{c}{60K} & \multicolumn{2}{c}{120K} & \multicolumn{2}{c}{128K} \\
retention & \IC{} & +\MLI{} & \IC{} & +\MLI{} & \IC{} & +\MLI{} & \IC{} & +\MLI{} \\
\midrule
none (base, tok/s) & \multicolumn{2}{c}{1760.5} & \multicolumn{2}{c}{494.7} & \multicolumn{2}{c}{316.3} & \multicolumn{2}{c}{314.2} \\
requests in flight & \multicolumn{2}{c}{64} & \multicolumn{2}{c}{13} & \multicolumn{2}{c}{6} & \multicolumn{2}{c}{6} \\
all (\MLI{} alone) & \multicolumn{2}{c}{-2.2} & \multicolumn{2}{c}{+4.3} & \multicolumn{2}{c}{+5.7} & \multicolumn{2}{c}{+9.8} \\
\midrule
$\ret{2}$ & +3.1 & +1.9 & +8.1 & +10.5 & +12.3 & +16.3 & +12.1 & +18.4 \\
$\ret{4}$ & +6.1 & +2.7 & +13.2 & +12.6 & +20.7 & +20.1 & +21.4 & +23.6 \\
$\ret{8}$ & +6.5 & +2.6 & +15.1 & +12.4 & +24.3 & +24.2 & +25.4 & +24.7 \\
$\ret{16}$ & +2.0 & +0.1 & +13.3 & +12.8 & +24.7 & +25.5 & +26.1 & +26.9 \\
$\ret{32}$ & +0.8 & +0.4 & +13.2 & +12.3 & +25.8 & +26.6 & +27.5 & +27.5 \\
\bottomrule
\end{tabular}
\vspace{4pt}
\begin{tabular}{lrrrrrrrr}
\toprule
\textbf{GLM-5} & \multicolumn{2}{c}{10K} & \multicolumn{2}{c}{60K} & \multicolumn{2}{c}{120K} & \multicolumn{2}{c}{195K} \\
retention & \IC{} & +\MLI{} & \IC{} & +\MLI{} & \IC{} & +\MLI{} & \IC{} & +\MLI{} \\
\midrule
none (base, tok/s) & \multicolumn{2}{c}{1129.6} & \multicolumn{2}{c}{322.2} & \multicolumn{2}{c}{158.2} & \multicolumn{2}{c}{103.0} \\
requests in flight & \multicolumn{2}{c}{43} & \multicolumn{2}{c}{7} & \multicolumn{2}{c}{3} & \multicolumn{2}{c}{2} \\
all (\MLI{} alone) & \multicolumn{2}{c}{+4.4} & \multicolumn{2}{c}{+7.1} & \multicolumn{2}{c}{+5.6} & \multicolumn{2}{c}{+11.3} \\
\midrule
$\ret{2}$ & +5.7 & +6.4 & +9.1 & +12.8 & +10.6 & +13.7 & +11.4 & +18.3 \\
$\ret{4}$ & +8.2 & +10.1 & +14.7 & +16.7 & +15.9 & +17.9 & +18.6 & +22.9 \\
$\ret{8}$ & +9.2 & +9.1 & +17.6 & +18.2 & +19.8 & +20.6 & +22.6 & +24.6 \\
$\ret{16}$ & +7.3 & +8.5 & +23.2 & +23.9 & +23.3 & +24.0 & +24.9 & +25.5 \\
$\ret{32}$ & +8.5 & +8.8 & +27.8 & +27.0 & +25.3 & +24.6 & +25.9 & +27.0 \\
\midrule
$\ret{2}$ searched & +6.4 & +6.2 & +9.5 & +11.1 & +10.7 & +14.1 & +11.6 & +17.2 \\
$\ret{4}$ searched & +9.4 & +9.6 & +14.6 & +14.4 & +16.6 & +18.1 & +17.9 & +21.4 \\
\bottomrule
\end{tabular}
\vspace{4pt}
\caption{Full-KV-cache decode throughput. Base row: aggregate tokens/s of the unmodified model with the request set sized to the KV cache (the number of requests decoding together is given per context); other cells: percentage change over base, positive = faster. Reference is the closing base run of the cell where the opening run was cold by more than $2\%$.}
\label{tab:decode-saturation}
\end{table}

\section{Prefill Latency}
\label{app:ttft}
\MLI{} acts on decode only; prefill runs the standard schedule, so the composition inherits IndexCache's prefill gain (from skipping the indexer on shared layers during prefill) and adds nothing to it; time-to-first-token matches IndexCache within measurement noise.

\section{Attention Data-Parallel Replication}
\label{app:dp}
\IC{} reports its numbers under attention data-parallelism. We replicate our cells under \texttt{--enable-dp-attention --dp-size 8} (attention and indexer of a request on one GPU, MoE tensor-parallel across eight) with the same gates. \Cref{tab:decode-per-request-dp} gives the per-request result; at single concurrency this layout leaves seven attention ranks idle, so base is slower than under tensor-parallel attention ($56.6$ vs $85.5$ tok/s at $128$K) and every indexer remedy hides or removes a smaller share of the step: \MLI{} $+3.9\%$, \IC{} $\ret{4}$ $+13.3\%$, the composition $+14.3\%$. The composition remains additive over \IC{} at every retention measured. \Cref{tab:decode-saturation-dp} gives full-KV-cache throughput under this layout.

\begin{table}[t]
\centering\small
\setlength{\tabcolsep}{4pt}
\renewcommand{\arraystretch}{0.9}
\begin{tabular}{lrr}
\toprule
\textbf{DeepSeek-V3.2} & \multicolumn{2}{c}{128K} \\
retention & \IC{} & +\MLI{} \\
\midrule
none (base, tok/s) & \multicolumn{2}{c}{56.6} \\
all (\MLI{} alone) & \multicolumn{2}{c}{+3.9} \\
\midrule
$\ret{2}$ & +8.9 & +10.5 \\
$\ret{4}$ & +13.3 & +14.3 \\
\bottomrule
\end{tabular}
\vspace{4pt}
\begin{tabular}{lrr}
\toprule
\textbf{GLM-5} & \multicolumn{2}{c}{195K} \\
retention & \IC{} & +\MLI{} \\
\midrule
none (base, tok/s) & \multicolumn{2}{c}{47.5} \\
all (\MLI{} alone) & \multicolumn{2}{c}{+5.8} \\
\midrule
$\ret{2}$ & +6.6 & +10.3 \\
$\ret{4}$ & +10.7 & +11.8 \\
\bottomrule
\end{tabular}
\vspace{4pt}
\caption{Replication under attention data-parallelism (\texttt{--enable-dp-attention --dp-size 8}, \IC{}'s parallelism): per-request decode throughput (tokens/s, single concurrency) relative to the unmodified model in the same cell; positive = faster. Decode throughput is the client-side output tokens / (latency $-$ TTFT), which agrees with the server-side metric within $0.5\%$ on the TP cells.}
\label{tab:decode-per-request-dp}
\end{table}

\begin{table}[t]
\centering\small
\setlength{\tabcolsep}{4pt}
\renewcommand{\arraystretch}{0.9}
\begin{tabular}{lrr}
\toprule
\textbf{DeepSeek-V3.2} & \multicolumn{2}{c}{128K} \\
retention & \IC{} & +\MLI{} \\
\midrule
none (base, tok/s) & \multicolumn{2}{c}{65.0} \\
requests in flight & \multicolumn{2}{c}{4} \\
all (\MLI{} alone) & \multicolumn{2}{c}{+2.3} \\
\midrule
$\ret{2}$ & +13.6 & +12.6 \\
$\ret{4}$ & +21.2 & +21.4 \\
$\ret{8}$ & +25.6 & +25.2 \\
$\ret{16}$ & +26.1 & +27.5 \\
$\ret{32}$ & +28.9 & +29.5 \\
\bottomrule
\end{tabular}
\vspace{4pt}
\begin{tabular}{lrr}
\toprule
\textbf{GLM-5} & \multicolumn{2}{c}{195K} \\
retention & \IC{} & +\MLI{} \\
\midrule
none (base, tok/s) & \multicolumn{2}{c}{12.3} \\
requests in flight & \multicolumn{2}{c}{1} \\
all (\MLI{} alone) & \multicolumn{2}{c}{+1.6} \\
\midrule
$\ret{2}$ & +11.8 & +10.7 \\
$\ret{4}$ & +14.8 & +15.6 \\
$\ret{8}$ & +18.3 & +18.3 \\
$\ret{16}$ & +20.8 & +19.9 \\
$\ret{32}$ & +21.5 & +20.6 \\
\bottomrule
\end{tabular}
\vspace{4pt}
\caption{Replication under attention data-parallelism: full-KV-cache aggregate decode throughput relative to the unmodified model; positive = faster.}
\label{tab:decode-saturation-dp}
\end{table}

\section{Searched Patterns}
\label{app:patterns}
We use IndexCache's published GLM-5 searched patterns~\citep{bai2026indexcache} ($40$ and $22$ full layers of $78$ at $\ret{2}$ and $\ret{4}$), applying \MLI{} to the same patterns.

\section{Full Quality Table and Frontier}
\label{app:quality-full}
\paragraph{LongBench~v2 protocol.} We follow the official evaluation script's non-CoT setting: its prompt, a $128$-token answer budget, and its extraction of ``The correct answer is (X)'', with greedy decoding where the script samples at temperature $0.1$. An answer without an extractable letter is scored as wrong, as in the script. Under this budget $6\%$ of the unmodified DeepSeek-V3.2 model's answers on the full $503$-question set are unparsed because the model begins a step-by-step explanation and is cut off; this applies to every arm alike and is part of the reported accuracy.

\paragraph{Serving protocol for reuse arms.} Every arm that retains a fraction of the indexers runs with each retained layer computing its own selection at prefill and copying it forward only at decode (\texttt{index\_prefill\_full\_indexer}), the configuration appropriate to models that ship indexer weights at every layer. Copying a stale selection at prefill as well corrupts the context encoding at aggressive retention and long context; that configuration is not reported.

\paragraph{Thinking mode.} LongBench~v2 is evaluated with each model's non-thinking format. A control on the first $50$ questions ($4096$-token budget) showed no lift for DeepSeek-V3.2 ($0.52$ thinking vs $0.62$ non-thinking, $9$ traces truncated) and, for GLM-5, reasoning that overran the budget on half the questions; the sweep was therefore run without thinking.

\begin{figure}[t]
\centering
\includegraphics[width=0.95\linewidth]{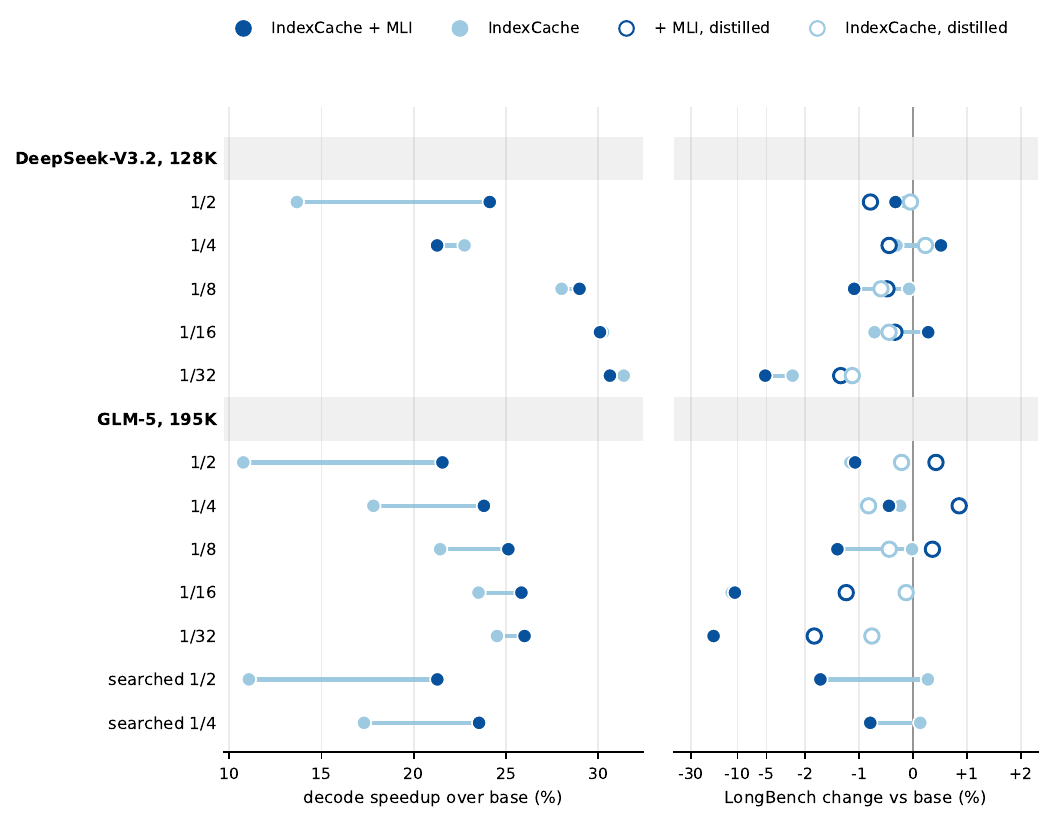}
\caption{What adding \MLI{} to IndexCache does at each retention, at the longest measured context of each model. Left: per-request decode speedup over the unmodified model; right: LongBench macro-F1 change (symmetric-log axis so the $\ret{32}$ collapse fits). Light: IndexCache alone; dark: IndexCache$+$\MLI{} at the same retention; hollow: distilled arms.}
\label{fig:frontier}
\end{figure}

\paragraph{Frontier.} \Cref{fig:frontier} plots per-request decode speedup against LongBench for IndexCache and IndexCache$+$\MLI{} at every retention, undistilled and distilled, at the longest context of each model: the composition sits up and to the left of IndexCache at matched retention, and one halving further in it avoids the quality collapse IndexCache falls into.
\begin{table}[t]
\centering\small
\begin{tabular}{lrrrrr}
\toprule
\textbf{DeepSeek-V3.2} & LongBench & NIAH & PPL & RULER & LongBench v2 \\
\midrule
base & 0.449 & 1.000 & 1.167 & 0.975 & 0.501 \\
\MLI{} & 0.448 (-0.0\%) & 1.00 & 1.176 (+0.8\%) & 0.954 (-2.1\%) & 0.499 (-0.4\%) \\
\MLI{} (distilled) & 0.447 (-0.4\%) & 1.00 & 1.203 (+3.1\%) & 0.929 (-4.7\%) & 0.479 (-4.4\%) \\
IC$_{1/2}$ & 0.448 (-0.2\%) & 1.00 & 1.166 (-0.1\%) & 0.949 (-2.7\%) & 0.505 (+0.8\%) \\
IC$_{1/2}$ (distilled) & 0.448 (-0.0\%) & 1.00 & 1.160 (-0.6\%) & 0.965 (-1.1\%) & 0.497 (-0.8\%) \\
IC$_{1/2}$+\MLI{} & 0.447 (-0.3\%) & 1.00 & 1.132 (-3.0\%) & 0.936 (-4.0\%) & 0.503 (+0.4\%) \\
IC$_{1/2}$+\MLI{} (distilled) & 0.445 (-0.8\%) & 1.00 & 1.142 (-2.1\%) & 0.939 (-3.7\%) & 0.501 (+0.0\%) \\
IC$_{1/4}$ & 0.447 (-0.3\%) & 1.00 & 1.193 (+2.3\%) & 0.966 (-0.9\%) & 0.493 (-1.6\%) \\
IC$_{1/4}$ (distilled) & 0.450 (+0.2\%) & 1.00 & 1.153 (-1.2\%) & 0.956 (-2.0\%) & 0.497 (-0.8\%) \\
IC$_{1/4}$+\MLI{} & 0.451 (+0.5\%) & 1.00 & 1.193 (+2.3\%) & 0.971 (-0.4\%) & 0.499 (-0.4\%) \\
IC$_{1/4}$+\MLI{} (distilled) & 0.447 (-0.4\%) & 1.00 & 1.162 (-0.4\%) & 0.936 (-4.0\%) & 0.497 (-0.8\%) \\
IC$_{1/8}$ & 0.448 (-0.1\%) & 1.00 & 1.193 (+2.2\%) & 0.968 (-0.7\%) & 0.499 (-0.4\%) \\
IC$_{1/8}$ (distilled) & 0.446 (-0.6\%) & 1.00 & 1.168 (+0.1\%) & 0.967 (-0.9\%) & 0.503 (+0.4\%) \\
IC$_{1/8}$+\MLI{} & 0.444 (-1.1\%) & 1.00 & 1.149 (-1.5\%) & 0.978 (+0.3\%) & 0.503 (+0.4\%) \\
IC$_{1/8}$+\MLI{} (distilled) & 0.447 (-0.5\%) & 1.00 & 1.111 (-4.8\%) & 0.949 (-2.7\%) & 0.499 (-0.4\%) \\
IC$_{1/16}$ & 0.445 (-0.7\%) & 1.00 & 1.175 (+0.7\%) & 0.961 (-1.4\%) & 0.473 (-5.6\%) \\
IC$_{1/16}$ (distilled) & 0.447 (-0.4\%) & 1.00 & 1.156 (-0.9\%) & 0.964 (-1.1\%) & 0.489 (-2.4\%) \\
IC$_{1/16}$+\MLI{} & 0.450 (+0.3\%) & 1.00 & 1.144 (-2.0\%) & 0.937 (-3.9\%) & 0.465 (-7.1\%) \\
IC$_{1/16}$+\MLI{} (distilled) & 0.447 (-0.3\%) & 1.00 & 1.142 (-2.1\%) & 0.958 (-1.7\%) & 0.493 (-1.6\%) \\
IC$_{1/32}$ & 0.437 (-2.7\%) & 0.11 & 1.140 (-2.3\%) & 0.891 (-8.6\%) & 0.352 (-29.8\%) \\
IC$_{1/32}$ (distilled) & 0.444 (-1.1\%) & 0.89 & 1.166 (-0.1\%) & 0.912 (-6.4\%) & 0.475 (-5.2\%) \\
IC$_{1/32}$+\MLI{} & 0.426 (-5.2\%) & 0.11 & 1.155 (-1.1\%) & 0.817 (-16.2\%) & 0.314 (-37.3\%) \\
IC$_{1/32}$+\MLI{} (distilled) & 0.443 (-1.3\%) & 0.22 & 1.141 (-2.2\%) & 0.776 (-20.4\%) & 0.467 (-6.7\%) \\
\bottomrule
\end{tabular}
\vspace{2pt}
\begin{tabular}{lrrrrr}
\toprule
\textbf{GLM-5} & LongBench & NIAH & PPL & RULER & LongBench v2 \\
\midrule
base & 0.453 & 1.000 & 1.163 & 0.941 & 0.477 \\
\MLI{} & 0.452 (-0.2\%) & 1.00 & 1.061 (-8.8\%) & 0.963 (+2.3\%) & 0.473 (-0.8\%) \\
\MLI{} (distilled) & 0.447 (-1.3\%) & 1.00 & 1.083 (-6.9\%) & 0.969 (+2.9\%) & 0.481 (+0.8\%) \\
IC$_{1/2}$ & 0.448 (-1.2\%) & 1.00 & 1.061 (-8.7\%) & 0.969 (+2.9\%) & 0.483 (+1.2\%) \\
IC$_{1/2}$ (distilled) & 0.452 (-0.2\%) & 1.00 & 1.082 (-7.0\%) & 0.960 (+2.0\%) & 0.491 (+2.9\%) \\
IC$_{1/2}$+\MLI{} & 0.448 (-1.1\%) & 1.00 & 1.081 (-7.1\%) & 0.961 (+2.1\%) & 0.479 (+0.4\%) \\
IC$_{1/2}$+\MLI{} (distilled) & 0.455 (+0.4\%) & 1.00 & 1.108 (-4.7\%) & 0.962 (+2.2\%) & 0.467 (-2.1\%) \\
IC$_{1/2}^{\mathrm{searched}}$ & 0.454 (+0.3\%) & 1.00 & 1.078 (-7.4\%) & 0.956 (+1.5\%) & 0.473 (-0.8\%) \\
IC$_{1/2}^{\mathrm{searched}}$+\MLI{} & 0.445 (-1.7\%) & 1.00 & 1.087 (-6.6\%) & 0.951 (+1.0\%) & 0.477 (+0.0\%) \\
IC$_{1/4}$ & 0.452 (-0.2\%) & 1.00 & 1.074 (-7.7\%) & 0.948 (+0.7\%) & 0.489 (+2.5\%) \\
IC$_{1/4}$ (distilled) & 0.449 (-0.8\%) & 1.00 & 1.105 (-5.0\%) & 0.950 (+0.9\%) & 0.473 (-0.8\%) \\
IC$_{1/4}$+\MLI{} & 0.451 (-0.4\%) & 1.00 & 1.132 (-2.6\%) & 0.957 (+1.6\%) & 0.475 (-0.4\%) \\
IC$_{1/4}$+\MLI{} (distilled) & 0.457 (+0.9\%) & 1.00 & 1.090 (-6.2\%) & 0.931 (-1.1\%) & 0.487 (+2.1\%) \\
IC$_{1/4}^{\mathrm{searched}}$ & 0.454 (+0.1\%) & 1.00 & 1.086 (-6.6\%) & 0.971 (+3.2\%) & 0.475 (-0.4\%) \\
IC$_{1/4}^{\mathrm{searched}}$+\MLI{} & 0.450 (-0.8\%) & 1.00 & 1.082 (-7.0\%) & 0.958 (+1.8\%) & 0.483 (+1.2\%) \\
IC$_{1/8}$ & 0.453 (-0.0\%) & 1.00 & 1.076 (-7.4\%) & 0.957 (+1.6\%) & 0.481 (+0.8\%) \\
IC$_{1/8}$ (distilled) & 0.451 (-0.4\%) & 1.00 & 1.094 (-5.9\%) & 0.952 (+1.2\%) & 0.477 (+0.0\%) \\
IC$_{1/8}$+\MLI{} & 0.447 (-1.4\%) & 1.00 & 1.084 (-6.8\%) & 0.929 (-1.3\%) & 0.469 (-1.7\%) \\
IC$_{1/8}$+\MLI{} (distilled) & 0.455 (+0.4\%) & 1.00 & 1.110 (-4.6\%) & 0.953 (+1.2\%) & 0.467 (-2.1\%) \\
IC$_{1/16}$ & 0.401 (-11.5\%) & 0.00 & 1.827 (+57.1\%) & 0.365 (-61.3\%) & 0.378 (-20.8\%) \\
IC$_{1/16}$ (distilled) & 0.453 (-0.1\%) & 1.00 & 1.105 (-5.0\%) & 0.871 (-7.4\%) & 0.485 (+1.7\%) \\
IC$_{1/16}$+\MLI{} & 0.405 (-10.6\%) & 0.00 & 1.894 (+62.8\%) & 0.541 (-42.6\%) & 0.388 (-18.8\%) \\
IC$_{1/16}$+\MLI{} (distilled) & 0.448 (-1.2\%) & 1.00 & 1.095 (-5.9\%) & 0.806 (-14.4\%) & 0.483 (+1.2\%) \\
IC$_{1/32}$ & 0.375 (-17.3\%) & 0.00 & 2.499 (+114.9\%) & 0.494 (-47.5\%) & 0.336 (-29.6\%) \\
IC$_{1/32}$ (distilled) & 0.450 (-0.8\%) & 1.00 & 1.093 (-6.0\%) & 0.806 (-14.4\%) & 0.461 (-3.3\%) \\
IC$_{1/32}$+\MLI{} & 0.373 (-17.6\%) & 0.00 & 2.501 (+115.0\%) & 0.653 (-30.6\%) & 0.342 (-28.3\%) \\
IC$_{1/32}$+\MLI{} (distilled) & 0.445 (-1.8\%) & 0.78 & 1.124 (-3.4\%) & 0.839 (-10.9\%) & 0.427 (-10.4\%) \\
\bottomrule
\end{tabular}
\vspace{2pt}
\caption{Quality vs the unmodified model (same job, same endpoints, greedy). LongBench: macro-F1 over 11 tasks; NIAH: worst score over haystacks up to 520K characters and both needle modes; PPL: perplexity of the model's own greedy continuation, geometric mean over 4K--64K inputs and five seeds (a deviation in either direction is drift from the unmodified model); RULER: mean over tasks and lengths 4K--128K; LongBench~v2: accuracy on the full 503-question set. Distilled rows use indexer-only span-aware distillation with the same recipe for both methods. A missing entry (--) is an evaluation whose server ran out of memory during prefill in two attempts.}
\label{tab:quality}
\end{table}

\begin{table}[t]
\centering\small
\setlength{\tabcolsep}{3.5pt}
\begin{tabular}{llrrrrrrr}
\toprule
 & & & \multicolumn{3}{c}{undistilled} & \multicolumn{3}{c}{distilled} \\
 & arm & decode & LongBench & RULER & LBv2 & LongBench & RULER & LBv2 \\
\midrule
\multicolumn{9}{l}{\textbf{DeepSeek-V3.2}, 128K} \\
 & \IC{} $\ret{4}$ & +22.7 & -0.3 & -0.9 & -1.6 & +0.2 & -2.0 & -0.8 \\
 & \IC{}$+$\MLI{} $\ret{2}$ & +29.8 & -0.3 & -4.0 & +0.4 & -0.8 & -3.7 & +0.0 \\
\addlinespace[2pt]
 & \IC{} $\ret{8}$ & +28.0 & -0.1 & -0.7 & -0.4 & -0.6 & -0.9 & +0.4 \\
 & \IC{}$+$\MLI{} $\ret{4}$ & +32.3 & +0.5 & -0.4 & -0.4 & -0.4 & -4.0 & -0.8 \\
\addlinespace[2pt]
 & \IC{} $\ret{16}$ & +30.2 & -0.7 & -1.4 & -5.6 & -0.4 & -1.1 & -2.4 \\
 & \IC{}$+$\MLI{} $\ret{8}$ & +33.7 & -1.1 & +0.3 & +0.4 & -0.5 & -2.7 & -0.4 \\
\midrule
\multicolumn{9}{l}{\textbf{GLM-5}, 195K} \\
 & \IC{} $\ret{4}$ & +17.7 & -0.2 & +0.7 & +2.5 & -0.8 & +0.9 & -0.8 \\
 & \IC{}$+$\MLI{} $\ret{2}$ & +21.4 & -1.1 & +2.1 & +0.4 & +0.4 & +2.2 & -2.1 \\
\addlinespace[2pt]
 & \IC{} $\ret{8}$ & +21.3 & -0.0 & +1.6 & +0.8 & -0.4 & +1.2 & +0.0 \\
 & \IC{}$+$\MLI{} $\ret{4}$ & +23.7 & -0.4 & +1.6 & -0.4 & +0.9 & -1.1 & +2.1 \\
\addlinespace[2pt]
 & \IC{} $\ret{16}$ & +23.4 & -11.5 & -61.3 & -20.8 & -0.1 & -7.4 & +1.7 \\
 & \IC{}$+$\MLI{} $\ret{8}$ & +25.0 & -1.4 & -1.3 & -1.7 & +0.4 & +1.2 & -2.1 \\
\bottomrule
\end{tabular}
\vspace{4pt}
\caption{The halving pair: \IC{} at $\ret{2r}$ and \IC{}$+$\MLI{} at $\ret{r}$, the latter with twice the fresh selections. Decode: per-request throughput change over base at the longest context of each model (\IC{}$+$\MLI{} at $\ret{r}$ is faster than \IC{} at $\ret{2r}$, not merely equal); quality: percentage change over base, positive = better.}
\label{tab:equal-speed}
\end{table}

\clearpage
\section{Confidence Intervals for the Quality Contrasts}
\label{app:quality-ci}
\Cref{tab:quality-ci-ds,tab:quality-ci-glm5} give paired bootstrap $95\%$ intervals for every quality contrast the text makes: each arm against the unmodified model, IndexCache$+$\MLI{} against IndexCache at equal retention, and IndexCache$+$\MLI{} at $\ret{r}$ against IndexCache at $\ret{2r}$ (equal speed). The resample is over examples and is applied to both arms of a contrast, so the interval is for the paired difference; intervals marked $^{*}$ exclude zero.
\begin{table}[p]
\centering\scriptsize
\setlength{\tabcolsep}{3pt}
\begin{tabular}{lrrr}
\toprule
\textbf{DeepSeek-V3.2} & LongBench & RULER & LongBench v2 \\
\midrule
IC$_{1/16}$ (distilled) $-$ base & -0.4 [-2.2, +1.3] & -1.1 [-3.3, +0.3] & -2.4 [-6.3, +1.6] \\
IC$_{1/2}$ (distilled) $-$ base & -0.0 [-1.8, +1.6] & -1.1 [-3.9, +1.6] & -0.8 [-4.4, +2.8] \\
IC$_{1/32}$ (distilled) $-$ base & -1.1 [-3.0, +0.7] & -6.4 [-9.9, -3.1]$^{*}$ & -5.2 [-9.5, -0.8]$^{*}$ \\
IC$_{1/4}$ (distilled) $-$ base & +0.2 [-1.7, +2.1] & -2.0 [-4.5, +0.4] & -0.8 [-4.0, +2.4] \\
IC$_{1/8}$ (distilled) $-$ base & -0.6 [-2.4, +1.2] & -0.9 [-2.5, +0.4] & +0.4 [-3.6, +4.4] \\
IC$_{1/16}$ $-$ base & -0.7 [-2.4, +0.9] & -1.4 [-3.6, +0.2] & -5.6 [-10.7, -0.4]$^{*}$ \\
IC$_{1/2}$ $-$ base & -0.2 [-2.0, +1.7] & -2.7 [-4.9, -0.5]$^{*}$ & +0.8 [-2.8, +4.4] \\
IC$_{1/32}$ $-$ base & -2.7 [-4.7, -0.8]$^{*}$ & -8.6 [-11.9, -5.5]$^{*}$ & -29.8 [-38.1, -21.4]$^{*}$ \\
IC$_{1/4}$ $-$ base & -0.3 [-1.9, +1.3] & -0.9 [-2.4, +0.1] & -1.6 [-4.8, +1.6] \\
IC$_{1/8}$ $-$ base & -0.1 [-2.0, +1.8] & -0.7 [-2.2, +0.4] & -0.4 [-3.6, +2.8] \\
IC$_{1/16}$+\MLI{} (distilled) $-$ base & -0.3 [-2.1, +1.4] & -1.7 [-3.8, -0.2]$^{*}$ & -1.6 [-5.6, +2.0] \\
IC$_{1/2}$+\MLI{} (distilled) $-$ base & -0.8 [-2.6, +1.0] & -3.7 [-6.4, -1.2]$^{*}$ & +0.0 [-4.0, +4.0] \\
IC$_{1/32}$+\MLI{} (distilled) $-$ base & -1.3 [-3.2, +0.5] & -20.4 [-23.2, -17.5]$^{*}$ & -6.7 [-11.5, -2.4]$^{*}$ \\
IC$_{1/4}$+\MLI{} (distilled) $-$ base & -0.4 [-2.1, +1.2] & -4.0 [-7.0, -1.0]$^{*}$ & -0.8 [-4.0, +2.4] \\
IC$_{1/8}$+\MLI{} (distilled) $-$ base & -0.5 [-2.2, +1.2] & -2.7 [-4.4, -1.1]$^{*}$ & -0.4 [-4.0, +3.2] \\
IC$_{1/16}$+\MLI{} $-$ base & +0.3 [-1.6, +2.1] & -3.9 [-6.3, -1.7]$^{*}$ & -7.1 [-12.3, -2.0]$^{*}$ \\
IC$_{1/2}$+\MLI{} $-$ base & -0.3 [-2.1, +1.4] & -4.0 [-7.1, -1.2]$^{*}$ & +0.4 [-2.0, +2.8] \\
IC$_{1/32}$+\MLI{} $-$ base & -5.2 [-7.5, -2.9]$^{*}$ & -16.2 [-20.1, -12.3]$^{*}$ & -37.3 [-46.0, -28.6]$^{*}$ \\
IC$_{1/4}$+\MLI{} $-$ base & +0.5 [-1.2, +2.2] & -0.4 [-3.2, +2.1] & -0.4 [-3.2, +2.4] \\
IC$_{1/8}$+\MLI{} $-$ base & -1.1 [-2.9, +0.6] & +0.3 [-0.1, +0.7] & +0.4 [-2.8, +3.6] \\
\MLI{} $-$ base & -0.0 [-1.1, +1.0] & -2.1 [-4.1, -0.2]$^{*}$ & -0.4 [-3.2, +2.8] \\
\MLI{} (distilled) $-$ base & -0.4 [-2.0, +1.3] & -4.7 [-7.3, -2.3]$^{*}$ & -4.4 [-9.5, +0.8] \\
equal retention $\ret{2}$: +\MLI{} $-$ \IC{} & -0.2 [-1.2, +0.9] & -1.4 [-4.1, +1.1] & -0.4 [-3.6, +2.8] \\
equal retention $\ret{2}$: +\MLI{} $-$ \IC{} (distilled) & -0.7 [-2.2, +0.7] & -2.6 [-5.1, -0.3]$^{*}$ & +0.8 [-3.2, +4.8] \\
equal retention $\ret{4}$: +\MLI{} $-$ \IC{} & +0.8 [-0.1, +1.8] & +0.5 [-2.4, +3.2] & +1.2 [-2.4, +4.8] \\
equal retention $\ret{4}$: +\MLI{} $-$ \IC{} (distilled) & -0.7 [-1.9, +0.6] & -2.0 [-4.3, +0.0] & +0.0 [-4.0, +4.0] \\
equal retention $\ret{8}$: +\MLI{} $-$ \IC{} & -1.0 [-2.2, +0.0] & +1.0 [+0.0, +2.5] & +0.8 [-2.0, +3.6] \\
equal retention $\ret{8}$: +\MLI{} $-$ \IC{} (distilled) & +0.1 [-1.1, +1.3] & -1.8 [-4.2, +0.7] & -0.8 [-4.4, +2.8] \\
equal retention $\ret{16}$: +\MLI{} $-$ \IC{} & +1.0 [-0.3, +2.3] & -2.5 [-4.6, -0.6]$^{*}$ & -1.6 [-6.7, +3.6] \\
equal retention $\ret{16}$: +\MLI{} $-$ \IC{} (distilled) & +0.1 [-1.0, +1.2] & -0.6 [-1.2, +0.0] & +0.8 [-2.8, +4.4] \\
equal retention $\ret{32}$: +\MLI{} $-$ \IC{} & -2.5 [-4.3, -0.8]$^{*}$ & -7.6 [-11.4, -3.4]$^{*}$ & -7.5 [-15.5, +0.0] \\
equal retention $\ret{32}$: +\MLI{} $-$ \IC{} (distilled) & -0.2 [-1.6, +1.2] & -14.0 [-16.7, -11.6]$^{*}$ & -1.6 [-5.6, +2.4] \\
equal speed: \IC{}+\MLI{} $\ret{2}$ $-$ \IC{} $\ret{4}$ & -0.0 [-1.1, +1.1] & -3.1 [-6.0, -0.7]$^{*}$ & +2.0 [-1.2, +5.2] \\
equal speed: \IC{}+\MLI{} $\ret{2}$ $-$ \IC{} $\ret{4}$ (distilled) & -1.0 [-2.5, +0.5] & -1.7 [-3.2, -0.4]$^{*}$ & +0.8 [-3.6, +5.2] \\
equal speed: \IC{}+\MLI{} $\ret{4}$ $-$ \IC{} $\ret{8}$ & +0.6 [-0.6, +1.9] & +0.4 [-2.5, +3.1] & +0.0 [-3.2, +3.2] \\
equal speed: \IC{}+\MLI{} $\ret{4}$ $-$ \IC{} $\ret{8}$ (distilled) & +0.2 [-1.0, +1.3] & -3.2 [-6.6, +0.2] & -1.2 [-4.8, +2.0] \\
equal speed: \IC{}+\MLI{} $\ret{8}$ $-$ \IC{} $\ret{16}$ & -0.4 [-1.5, +0.7] & +1.7 [+0.2, +3.8]$^{*}$ & +6.0 [+0.8, +11.1]$^{*}$ \\
equal speed: \IC{}+\MLI{} $\ret{8}$ $-$ \IC{} $\ret{16}$ (distilled) & -0.0 [-1.1, +1.0] & -1.5 [-3.4, +0.2] & +2.0 [-1.6, +5.6] \\
equal speed: \IC{}+\MLI{} $\ret{16}$ $-$ \IC{} $\ret{32}$ & +3.0 [+1.3, +4.7]$^{*}$ & +4.7 [+1.5, +7.9]$^{*}$ & +22.6 [+14.3, +31.3]$^{*}$ \\
equal speed: \IC{}+\MLI{} $\ret{16}$ $-$ \IC{} $\ret{32}$ (distilled) & +0.8 [-0.4, +2.1] & +4.7 [+1.7, +7.4]$^{*}$ & +3.6 [-1.2, +8.3] \\
\bottomrule
\end{tabular}
\caption{DeepSeek-V3.2. Paired bootstrap $95\%$ intervals for the contrasts in the text (percentage change over the unmodified model's score; $10{,}000$ resamples of examples, the same resample applied to both arms; LongBench resampled within each of its 11 tasks and RULER within each of its 30 task$\times$length cells, both macro-averaged as in \Cref{tab:quality}). $^{*}$: interval excludes zero. Examples per arm: LongBench $1{,}100$, RULER $234$, LongBench~v2 as stated in \Cref{tab:quality}.}
\label{tab:quality-ci-ds}
\end{table}
\begin{table}[p]
\centering\scriptsize
\setlength{\tabcolsep}{3pt}
\begin{tabular}{lrrr}
\toprule
\textbf{GLM-5} & LongBench & RULER & LongBench v2 \\
\midrule
IC$_{1/16}$ (distilled) $-$ base & -0.1 [-2.2, +2.0] & -7.4 [-10.2, -4.7]$^{*}$ & +1.7 [-2.9, +6.2] \\
IC$_{1/2}$ (distilled) $-$ base & -0.2 [-2.2, +1.8] & +2.0 [+0.4, +3.6]$^{*}$ & +2.9 [-1.7, +7.9] \\
IC$_{1/32}$ (distilled) $-$ base & -0.8 [-3.0, +1.5] & -14.4 [-18.6, -10.2]$^{*}$ & -3.3 [-9.6, +2.5] \\
IC$_{1/4}$ (distilled) $-$ base & -0.8 [-2.9, +1.1] & +0.9 [-2.4, +4.1] & -0.8 [-5.0, +3.3] \\
IC$_{1/8}$ (distilled) $-$ base & -0.4 [-2.4, +1.5] & +1.2 [-1.4, +3.7] & +0.0 [-4.6, +4.6] \\
IC$_{1/16}$ $-$ base & -11.5 [-14.4, -8.5]$^{*}$ & -61.3 [-64.9, -57.7]$^{*}$ & -20.8 [-27.9, -13.8]$^{*}$ \\
IC$_{1/2}$ $-$ base & -1.2 [-3.2, +0.9] & +2.9 [+0.9, +4.9]$^{*}$ & +1.3 [-2.1, +4.6] \\
IC$_{1/32}$ $-$ base & -17.3 [-20.7, -13.9]$^{*}$ & -47.5 [-53.0, -42.2]$^{*}$ & -29.6 [-36.7, -22.1]$^{*}$ \\
IC$_{1/4}$ $-$ base & -0.2 [-2.4, +1.9] & +0.7 [-1.3, +2.6] & +2.5 [-1.7, +6.7] \\
IC$_{1/8}$ $-$ base & -0.0 [-2.1, +2.1] & +1.6 [+0.2, +3.1]$^{*}$ & +0.8 [-2.9, +4.6] \\
IC$_{1/16}$+\MLI{} (distilled) $-$ base & -1.2 [-3.6, +1.1] & -14.4 [-17.7, -11.2]$^{*}$ & +1.3 [-3.8, +6.2] \\
IC$_{1/2}$+\MLI{} (distilled) $-$ base & +0.4 [-1.5, +2.4] & +2.2 [-1.8, +6.1] & -2.1 [-6.7, +2.5] \\
IC$_{1/32}$+\MLI{} (distilled) $-$ base & -1.8 [-4.1, +0.4] & -10.9 [-15.1, -6.9]$^{*}$ & -10.4 [-17.5, -3.8]$^{*}$ \\
IC$_{1/4}$+\MLI{} (distilled) $-$ base & +0.9 [-1.2, +2.9] & -1.1 [-3.5, +1.4] & +2.1 [-2.9, +7.1] \\
IC$_{1/8}$+\MLI{} (distilled) $-$ base & +0.4 [-1.8, +2.6] & +1.2 [-0.3, +2.7] & -2.1 [-7.1, +2.9] \\
IC$_{1/16}$+\MLI{} $-$ base & -10.6 [-13.5, -7.7]$^{*}$ & -42.6 [-47.1, -38.0]$^{*}$ & -18.7 [-25.4, -12.1]$^{*}$ \\
IC$_{1/2}$+\MLI{} $-$ base & -1.1 [-3.1, +0.9] & +2.1 [+0.7, +3.4]$^{*}$ & +0.4 [-3.8, +4.6] \\
IC$_{1/32}$+\MLI{} $-$ base & -17.6 [-21.0, -14.2]$^{*}$ & -30.6 [-35.5, -25.8]$^{*}$ & -28.3 [-35.4, -21.3]$^{*}$ \\
IC$_{1/4}$+\MLI{} $-$ base & -0.4 [-2.3, +1.3] & +1.6 [-0.6, +3.7] & -0.4 [-4.6, +4.2] \\
IC$_{1/8}$+\MLI{} $-$ base & -1.4 [-3.7, +0.9] & -1.3 [-4.1, +1.3] & -1.7 [-5.8, +2.5] \\
IC$_{1/2}^{\mathrm{searched}}$+\MLI{} $-$ base & -1.7 [-3.7, +0.2] & +1.0 [-1.4, +3.2] & +0.0 [-3.8, +3.8] \\
IC$_{1/4}^{\mathrm{searched}}$+\MLI{} $-$ base & -0.8 [-2.9, +1.3] & +1.8 [-0.6, +4.0] & +1.3 [-2.5, +5.4] \\
IC$_{1/2}^{\mathrm{searched}}$ $-$ base & +0.3 [-1.9, +2.5] & +1.5 [-1.4, +4.7] & -0.8 [-5.0, +3.3] \\
IC$_{1/4}^{\mathrm{searched}}$ $-$ base & +0.1 [-1.9, +2.3] & +3.2 [+1.6, +4.6]$^{*}$ & -0.4 [-4.6, +3.8] \\
\MLI{} $-$ base & -0.2 [-2.2, +1.7] & +2.3 [+0.4, +4.3]$^{*}$ & -0.8 [-5.0, +3.3] \\
\MLI{} (distilled) $-$ base & -1.3 [-3.5, +0.9] & +2.9 [+0.6, +5.2]$^{*}$ & +0.8 [-4.2, +5.8] \\
equal retention $\ret{2}$: +\MLI{} $-$ \IC{} & +0.1 [-2.1, +2.2] & -0.8 [-2.6, +0.8] & -0.8 [-5.0, +3.3] \\
equal retention $\ret{2}$: +\MLI{} $-$ \IC{} (distilled) & +0.6 [-1.3, +2.6] & +0.2 [-3.3, +4.0] & -5.0 [-9.2, -1.2]$^{*}$ \\
equal retention $\ret{4}$: +\MLI{} $-$ \IC{} & -0.2 [-2.3, +1.8] & +0.9 [-1.8, +3.3] & -2.9 [-7.1, +1.3] \\
equal retention $\ret{4}$: +\MLI{} $-$ \IC{} (distilled) & +1.7 [-0.6, +4.0] & -2.0 [-4.7, +0.8] & +2.9 [-1.7, +7.5] \\
equal retention $\ret{8}$: +\MLI{} $-$ \IC{} & -1.4 [-3.6, +0.7] & -3.0 [-5.2, -0.7]$^{*}$ & -2.5 [-6.2, +1.2] \\
equal retention $\ret{8}$: +\MLI{} $-$ \IC{} (distilled) & +0.8 [-1.5, +3.1] & +0.0 [-2.5, +2.6] & -2.1 [-6.7, +2.5] \\
equal retention $\ret{16}$: +\MLI{} $-$ \IC{} & +0.8 [-1.6, +3.3] & +18.7 [+14.4, +23.0]$^{*}$ & +2.1 [-3.8, +7.9] \\
equal retention $\ret{16}$: +\MLI{} $-$ \IC{} (distilled) & -1.1 [-3.5, +1.2] & -7.0 [-10.4, -3.5]$^{*}$ & -0.4 [-5.0, +4.2] \\
equal retention $\ret{32}$: +\MLI{} $-$ \IC{} & -0.3 [-2.6, +2.0] & +16.9 [+12.0, +21.8]$^{*}$ & +1.2 [-3.3, +5.4] \\
equal retention $\ret{32}$: +\MLI{} $-$ \IC{} (distilled) & -1.1 [-3.5, +1.4] & +3.5 [+0.0, +6.9]$^{*}$ & -7.1 [-13.8, -0.8]$^{*}$ \\
equal speed: \IC{}+\MLI{} $\ret{2}$ $-$ \IC{} $\ret{4}$ & -0.8 [-3.1, +1.4] & +1.4 [-0.7, +3.5] & -2.1 [-6.7, +2.1] \\
equal speed: \IC{}+\MLI{} $\ret{2}$ $-$ \IC{} $\ret{4}$ (distilled) & +1.2 [-0.9, +3.4] & +1.3 [-0.4, +3.1] & -1.2 [-5.8, +3.3] \\
equal speed: \IC{}+\MLI{} $\ret{4}$ $-$ \IC{} $\ret{8}$ & -0.4 [-2.8, +2.0] & +0.0 [-2.4, +2.0] & -1.2 [-5.4, +2.5] \\
equal speed: \IC{}+\MLI{} $\ret{4}$ $-$ \IC{} $\ret{8}$ (distilled) & +1.3 [-0.8, +3.4] & -2.3 [-5.3, +0.7] & +2.1 [-2.9, +7.1] \\
equal speed: \IC{}+\MLI{} $\ret{8}$ $-$ \IC{} $\ret{16}$ & +10.1 [+6.8, +13.3]$^{*}$ & +59.9 [+55.8, +63.8]$^{*}$ & +19.2 [+12.5, +26.2]$^{*}$ \\
equal speed: \IC{}+\MLI{} $\ret{8}$ $-$ \IC{} $\ret{16}$ (distilled) & +0.5 [-1.7, +2.7] & +8.7 [+6.0, +11.3]$^{*}$ & -3.7 [-8.3, +1.2] \\
equal speed: \IC{}+\MLI{} $\ret{16}$ $-$ \IC{} $\ret{32}$ & +6.7 [+4.2, +9.4]$^{*}$ & +5.0 [-0.4, +10.4] & +10.8 [+5.0, +16.7]$^{*}$ \\
equal speed: \IC{}+\MLI{} $\ret{16}$ $-$ \IC{} $\ret{32}$ (distilled) & -0.5 [-2.7, +1.7] & -0.0 [-3.7, +3.7] & +4.6 [-2.1, +11.2] \\
\bottomrule
\end{tabular}
\caption{GLM-5. Paired bootstrap $95\%$ intervals for the contrasts in the text (percentage change over the unmodified model's score; $10{,}000$ resamples of examples, the same resample applied to both arms; LongBench resampled within each of its 11 tasks and RULER within each of its 30 task$\times$length cells, both macro-averaged as in \Cref{tab:quality}). $^{*}$: interval excludes zero. Examples per arm: LongBench $1{,}100$, RULER $234$, LongBench~v2 as stated in \Cref{tab:quality}.}
\label{tab:quality-ci-glm5}
\end{table}

\section{Measurement Noise and Compute}
\label{app:noise}
Three sources of variation bound every speed number. \emph{Within a boot}: repeating the per-request measurement five times on one warm server gives a standard error of $0.01$ to $0.03$ tok/s per arm, so the equal-retention gaps carry $95\%$ intervals of about $\pm0.1$ percentage point (DeepSeek-V3.2 at $128$K: $\ret{4}$ $+4.70\pm0.11$, $\ret{8}$ $+2.36\pm0.14$; GLM-5 at $195$K: $\ret{4}$ $+4.12\pm0.06$, $\ret{8}$ $+2.61\pm0.06$). \emph{Between boots in a cell}: the closing run of the unmodified model is within $0.4\%$ of the opening one in every certified cell. \emph{Between cells}: each replicate cell and its sweep cell, run on different days and nodes, differ by $0.8\%$ (DeepSeek-V3.2) and $0.1\%$ (GLM-5) on the unmodified model and by up to $0.5$ and $0.2$ percentage point on the equal-retention gaps. The between-cell figure is the one to read against the tables; the equal-retention margins at $\ret{2}$ to $\ret{8}$ exceed it several times over.

All experiments ran on nodes of eight H200 GPUs (141~GB each) on an internal Slurm cluster. The reported speed sweep ($20$ cells plus $4$ data-parallel and $2$ replicate-CI cells, at $2$ to $4$ node-hours each) is about $90$ node-hours; the quality matrix ($48$ arms) about $75$; and distillation ($18$ runs) about $70$. Including the fused-kernel realization and cells voided by the gates, the project used roughly three times the reported compute.

\section{Distillation Recipe}
\label{app:distill}
Let $\mathcal{C}(f)$ be the consumers of full layer $f$ (\Cref{sec:method-distill}). For a query row $t$ let $I^{c}_{t,\cdot}$ be the unmodified model's indexer scores at consumer $c$ and $\hat I^{f}_{t,\cdot}$ the scores of the indexer being trained at $f$. The loss is
\begin{equation}
  \mathcal{L}_f \;=\; \frac{1}{|\mathcal{C}(f)|}\sum_{c\in\mathcal{C}(f)} \Big[\, \lambda\,\mathrm{KL}\!\big(\mathrm{softmax}(I^{c}_{t,\cdot})\,\big\|\,\mathrm{softmax}(\hat I^{f}_{t,\cdot})\big) \;+\; (1-\lambda)\, \mathcal{L}_{\mathrm{PL}}\big(\pi^{c}_{t}\,;\,\hat I^{f}_{t,\cdot}\big) \Big],
  \label{eq:distill}
\end{equation}
where $\mathcal{L}_{\mathrm{PL}}$ is the negative Plackett--Luce log-likelihood of the teacher's ranking $\pi^c_t$ of its top-$k$ tokens under the student's scores, with $\lambda{=}0.5$ and $k{=}2048$. The KL term alone leaves the ranking near the selection boundary free; the listwise term fixes it there.

Only the lightning indexers of the retained (full) layers are trained; the backbone, attention and the shared layers' behavior are frozen. Teacher signal: the unmodified model's indexer logits at the consumer layers, captured by forward hooks in a frozen forward over the same input. For full layer $\ell$ with next full layer $\ell'$, the consumers are layers $\ell{+}1,\dots,\ell'{-}1$ under \MLI{} (each attends $\Isel^{(\ell)}$ one full layer late) and $\ell,\dots,\ell'{-}1$ under \IC{} (the retained layer also serves itself); the student at $\ell$ is trained against every consumer in its span, with the same per-consumer weight. Objective: $0.5\,\mathrm{KL}$ between the consumer's softmaxed teacher scores and the student's, plus $0.5$ listwise Plackett--Luce likelihood of the teacher's top-$2048$ ordering under the student (the \emph{hybrid} objective), $512$ query rows sampled per step, causal masking. Data: GovReport documents packed into $8$ streams of $32{,}768$ tokens (DeepSeek-V3.2) or $16{,}384$ (GLM-5). Optimization: $300$ steps, AdamW, learning rate $10^{-3}$, no weight decay, gradient clipping at $1.0$, seed $0$, two H200 nodes per run. One run per (model, retention, span rule); the weights load at serving time through \texttt{index\_distilled\_weights} and change nothing else about the server. The identical recipe, data and budget are used for the \IC{} distilled arms, differing only in the span rule.

\end{document}